\documentclass[lettersize,journal]{IEEEtran}
\usepackage{amsmath,amsfonts}
\usepackage{algorithmic}
\usepackage{algorithm}
\usepackage{array}
\usepackage{textcomp}
\usepackage{stfloats}
\usepackage{url}
\usepackage{verbatim}
\usepackage{graphicx}
\usepackage{cite}

\usepackage{subfigure}
\usepackage{dutchcal} 
\usepackage{amssymb}

\usepackage{enumitem}

\usepackage{booktabs}

\usepackage{multirow}

\usepackage{makecell}

\usepackage{newfloat}
\usepackage{listings}

\begin{document}

\title{\vspace*{\fill} {\scriptsize This work has been accepted by the IEEE for publication. The copyright has been transferred. DOI: 10.1109/TMM.2023.3325965.} \\ Multi-task Paired Masking with Alignment Modeling for Medical Vision-Language Pre-training \\ \vspace*{\fill}}

\author{Ke Zhang, Yan Yang, Jun Yu,~\IEEEmembership{Senior Member,~IEEE,} Hanliang Jiang, Jianping Fan, Qingming Huang,~\IEEEmembership{Fellow,~IEEE,} and Weidong Han
\thanks{
This work was supported by the National Natural Science Foundation of China under Grants 62125201, 62020106007, 62176230. (Corresponding authors: Jun Yu and Weidong Han.) \par
K. Zhang, Y. Yang, and J. Yu are with the Key Laboratory of Complex Systems Modeling and Simulation, School of Computer Science and Technology, Hangzhou Dianzi University, Hangzhou, 310018, China (e-mail: ke.zhang@hdu.edu.cn; yangyan@hdu.edu.cn; yujun@hdu.edu.cn).  \par
H. Jiang is with Regional Medical Center for National Institute of Respiratory Diseases, Sir Run Run Shaw Hospital, College of Medicine, Zhejiang University, Hangzhou, 310016, China (e-mail: aock@zju.edu.cn).\par
J. Fan is with AI Lab at Lenovo Research, 100094, China (e-mail: jfan1@Lenovo.com). \par
Q. Huang is with the School of Computer Science and Technology, University of Chinese Academy of Sciences, Beijing, 101408, China (e-mail: qmhuang@ucas.ac.cn).\par
W. Han is affiliated with the Department of Medical Oncology at Sir Run Run Shaw Hospital, College of Medicine, Zhejiang University in Hangzhou, China. Additionally, he is a professor in the College of Mathematical Medicine at Zhejiang Normal University in Jinhua, China (e-mail: hanwd@zju.edu.cn).
}
}



\maketitle

\begin{abstract}
In recent years, the growing demand for medical imaging diagnosis has placed a significant burden on radiologists. As a solution, Medical Vision-Language Pre-training (Med-VLP) methods have been proposed to learn universal representations from medical images and reports, benefiting downstream tasks without requiring fine-grained annotations. However, existing methods have overlooked the importance of cross-modal alignment in joint image-text reconstruction, resulting in insufficient cross-modal interaction. To address this limitation, we propose a unified Med-VLP framework based on Multi-task Paired Masking with Alignment (MPMA) to integrate the cross-modal alignment task into the joint image-text reconstruction framework to achieve more comprehensive cross-modal interaction, while a Global and Local Alignment (GLA) module is designed to assist self-supervised paradigm in obtaining semantic representations with rich domain knowledge. Furthermore, we introduce a Memory-Augmented Cross-Modal Fusion (MA-CMF) module to fully integrate visual information to assist report reconstruction and fuse the multi-modal representations adequately. Experimental results demonstrate that the proposed unified approach outperforms previous methods in all downstream tasks, including uni-modal, cross-modal, and multi-modal tasks.

\end{abstract}

\begin{IEEEkeywords}
Medical vision-language pre-training, joint image-text reconstruction, cross-modal alignment.
\end{IEEEkeywords}

\section{Introduction}
\IEEEPARstart{A}{dvances} in medical imaging technology have significantly improved medical practice, but the growing number of medical images has increased the burden on radiologists. To address this issue, deep learning techniques have been widely used for automatic medical image analysis. However, these approaches require large-scale medical datasets with time-consuming and labor-intensive manual labeling, making them challenging to implement in practice. To overcome this challenge, Medical Vision-Language Pre-training (Med-VLP) has been proposed to leverage information-rich medical reports as supervised signals for learning generic representations from large-scale medical image-text data. Existing Med-VLP approaches aim to learn generic representations that can be utilized for various medical visual and textual tasks (e.g., medical visual question answering, medical image classification, and medical report generation), alleviating the problem of insufficient data volume in downstream tasks. In contrast to supervised learning, which relies on manual object-level annotations, Med-VLP leverages unlabeled medical image-text data to learn an aligned mapping and semantic representation from the latent space of raw text and images \cite{wang2022medclip}. Despite the advantages of self-supervised learning, Med-VLP still faces significant challenges due to the small publicly available datasets in radiology (e.g., only a few hundred thousand image-text pairs in \cite{mimic}) compared to the millions of data available in natural domains (e.g., \cite{CLIP-ViT-B} collected 400 million natural image-text pairs for self-supervised training).  \par
Current studies in Med-VLP mainly focus on two aspects: One is to improve the pre-training of cross-modal alignment based on ConVIRT \cite{ConVIRT}, and the other is the recently proposed joint image-text reconstruction for self-supervised pre-training, which applies the currently popular MAE \cite{MAE} method. For the latter method, Chen \textit{et al.} \cite{M3AE} implements unified multi-modal transformers with normal cross attention while Zhou \textit{et al.} \cite{MRM} utilizes visual information to assist in report reconstruction with only a simple Global Average Pooling (GAP) to perform cross-modal fusion. However, we find that these self-supervised reconstruction methods pay more attention to the invariance of feature representations within a single modality, neglecting the correlation between multiple modalities. This motivates us to combine the alignment-based and reconstruction-based methods and focus more on ensuring the relation between the modalities, incorporating both inter- and intra-modality tasks to promote each other. Moreover, the current joint reconstruction methods only use simple GAP to fuse images and reports, severely limiting the auxiliary impact of visual information on report reconstruction. Encouragingly, downstream benchmark tests have shown that building specific optimized cross-modal fusion strategies can help improve report reconstruction performance. \par
To address these limitations, we propose a unified Med-VLP framework based on Multi-task Paired Masking with Alignment (MPMA). Our MPMA integrates the cross-modal alignment task into the existing joint image-text reconstruction framework to achieve more comprehensive cross-modal interaction. To the best of our knowledge, our proposed framework is the first attempt to integrate both of the above in the medical domain. To achieve this, we introduce a global and local alignment (GLA) module for cross-modal alignment tasks to assist self-supervised paradigm in obtaining semantic representations with rich domain knowledge. Furthermore, in order to fully integrate visual information to assist report reconstruction, we propose a Memory-Augmented Cross-Modal Fusion (MA-CMF) module to fuse the multi-modal representations adequately. We perform pre-training on two large-scale medical datasets, namely MIMIC-CXR \cite{mimic} and ROCO \cite{ROCO}. To verify the effectiveness and generalization of our pre-training framework, we construct a medical vision-language understanding benchmark that includes uni-modal, cross-modal, and multi-modal tasks, corresponding to medical image classification, medical report generation, and medical visual question answering. \par
Our contributions are summarized as follows: \par

\begin{itemize}
    \item We propose a novel Multi-task Paired Masking with Alignment (MPMA) framework for Med-VLP, which aims to integrate joint image-text reconstruction tasks and cross-modal alignment tasks to achieve more comprehensive cross-modal interaction. 
    \item To achieve more effective cross-modal retrieval capabilities, we design a global and local alignment (GLA) module to assist self-supervised paradigm in obtaining more sufficient contextual representation.
    \item In order to facilitate the integration of visual information to assist report reconstruction, a novel Memory-Augmented Cross-modal Fusion (MA-CMF) module is proposed to fuse multi-modal information adequately.
    \item Our proposed MPMA framework can unify three different types of tasks (i.e., uni-modal, cross-modal, and multi-modal tasks). Experimental results show that our approach outperforms previous methods on all downstream tasks over six datasets, showcasing strong model generalization capabilities.
\end{itemize}

\section{Related Works}
\subsection{Medical Vision-Language Pre-training}
Currently, medical vision-language pre-training can be divided into two categories: report-supervised cross-modal alignment pre-training and reconstruction-based self-supervised pre-training. The former was first proposed by ConVIRT \cite{ConVIRT}, which employed radiological free-text reports to guide visual representation learning through bidirectional contrastive learning between two modalities. This approach was more consistent with the high inter-class similarity characteristics of medical images compared to the previous strategy of using ImageNet pre-training weights. However, bidirectional comparisons that focus solely on global features overlook the importance of local features. To address this issue, GLoRIA \cite{gloria} proposed a more effective supervision method by comparing the local similarity between image sub-regions and words from paired reports. Moreover, MGCA \cite{MGCA} believed that applying only \cite{gloria} was insufficient to mine enough semantic information at the instance and region levels, and thus proposed a disease-level alignment paradigm to enforce cross-modal cluster assignment consistency at a higher semantic level. On the basis of \cite{gloria}, BioViL \cite{BioviL} proposed a new language model, CXR-BERT, for natural language reasoning in radiology to improve textual modeling, and a new dataset MS-CXR was released, which included partially aligned image bounding boxes with phrase annotations. Inspired by \cite{ConVIRT}, REFERS \cite{REFERS} proposed to use multiple image views for each patient to learn joint representations and enhance the original visual information, but the dataset selection process resulted in a smaller dataset when medical data was already scarce. MedKLIP \cite{medklip} did not modify the pre-training dataset like \cite{REFERS}, nor did it directly perform simple feature extraction on medical reports. Instead, it proposed a triplet extraction module to extract the medical-related information from reports, using the previously extracted medical information to query from a knowledge base with entity translation. Med-CLIP \cite{wang2022medclip} endeavors to enhance cross-modal alignment through the mitigation of false negatives using a soft semantic matching loss. However, it encounters challenges related to the extraction of inaccurate medical tags and phrases, occasionally resulting in the generation of erroneous supervisory signals. The method based on cross-modal alignment is not only applied in medical and natural images but also widely used in video content \cite{cheng2017video1, cheng2017video2, cheng2016video3, nguyen2017vireo, cheng2017selection}.\par
Reconstruction-based self-supervised pre-training methods were mainly inspired by MAE \cite{MAE} and the self-supervised pre-text task in BERT \cite{Bert}, which used joint reconstruction of images and reports for representation learning. M3AE \cite{M3AE} learned cross-modal domain knowledge by reconstructing missing pixels and tokens from randomly masked images and texts, but did not fully exploit the complementarity between image and report completion. To tackle this problem, MRM \cite{MRM} proposed to directly integrate the cross-modal visual information processed by GAP during report reconstruction to learn knowledge-enhanced semantic representations, but neither M3AE nor MRM fully explored the integration of visual features. To address this limitation, we propose a Memory-Augmented Cross-Modal Fusion (MA-CMF) module to fully integrate visual features in the report reconstruction process. Additionally, we propose integrating cross-modal alignment pre-training into joint reconstruction pre-training processes and designing a global and local alignment (GLA) module to assist the self-supervised paradigm in obtaining semantic representations with rich domain knowledge.

\subsection{Medical Report Generation}
Medical report generation can be divided into two categories based on the development timeline: Recurrent Neural Network (RNN) based generative model and Transformer based generative model \cite{yangyan_jointembed}. With reference to the latest progress in computer vision \cite{yang2021adaptive} and machine translation \cite{cai2021chestxraybert}, Vinyals \textit{et al.} \cite{show-tell} proposed an end-to-end encoder-decoder architecture for image captioning, which served as the foundation for subsequent research. Rennie \textit{et al.} \cite{Att2in} introduced reward for self-critical based on \cite{show-tell}, which better encouraged consistency in training and testing performance. Lu \textit{et al.} \cite{AdaAtt} proposed an adaptive attention model with a visual sentinel that enabled the decoder to have different attention strategies for different types of words. \par
With the introduction of the Transformer \cite{transformer}, the RNN-based decoder was gradually replaced by the Transformer architecture due to its superior ability to encode and decode long paragraphs. Chen \textit{et al.} \cite{R2Gen} proposed a memory-driven transformer with relational memory to record key information of the generation process. R2GenCMN \cite{R2GenCMN} focused on cross-modal mapping that was previously overlooked, and designed a shared memory to record the alignment between images and texts, facilitating the interaction based on \cite{R2Gen}. PPKED \cite{PPKED} integrated report retrieval and medical prior knowledge to generate reports, simulating the working mode of radiologists. Unlike previous work that directly generated reports, AlignTrans \cite{aligntransformer} first predicted disease labels from input images and then learned multi-granularity visual features by hierarchically aligning visual regions and disease labels. Based on the cross-modal memory in \cite{R2GenCMN}, Qin \textit{et al.} \cite{CMM+RL} proposed to apply reinforcement learning to avoid overlapping valid dependencies due to the lack of direct cross-modal alignment annotations. RAMT \cite{RAMT-U} focused on the issue of data scarcity, taking the lead in applying semi-supervised learning to report generation tasks and proposing a relation-aware mean teacher framework to capture each potential pathological change. Moreover, Wang \textit{et al.} \cite{Multi-Criteria} enforced better cross-modal alignment and multi-label classification jointly using a pure transformer. Clinical-BERT \cite{clinical-bert} specially designed multiple sub-training tasks for report generation with a vision-language pre-training model. Our work is different from \cite{clinical-bert} in that our pre-training framework with extra image reconstruction is generic rather than specifically designed for a particular task, and we validate its effectiveness on more downstream tasks.

\subsection{Medical Visual Question Answering}
Medical Visual Question Answering (Med-VQA) has become a popular research field in recent years, thanks to the introduction of related large-scale datasets. The existing methods mainly focus on enhancing multi-modal representation and reasoning, including attention mechanisms, bilinear pooling, meta-learning, and data augmentation, etc. \par
SAN \cite{SAN} was the first to use a stacked attention model for multi-step reasoning to identify image regions associated with the questions while BAN \cite{BAN} utilized low-rank bilinear pooling to reduce the rank of the weight matrix. However, directly applying the above methods from the natural domain to the medical domain can lead to overfitting. To tackle this problem, Nguyen \textit{et al.} \cite{SAN-MEVF(BAN-MEVF)} applied unsupervised Denoising Auto Encoder (DAE) and supervised meta-learning to utilize a large amount of unlabeled data and learn meta weights adaptable to VQA problems. Zhan \textit{et al.} \cite{BAN-MEVF+CR} focused on question-conditioned reasoning and proposed a conditional reasoning-based framework to decouple open-ended and closed-ended questions.    \par
Some work focuses on applying data augmentation techniques to improve performance. Kafle \textit{et al.} \cite{BAN-MEVF+DAVQA} was the first to use data augmentation techniques, attempting to generate new questions using LSTM \cite{lstm} and template-based approaches. Different from the former, Tang \textit{et al.} \cite{BAN-MEVF+SEADA} did not directly make changes to images or questions, but instead generated adversarial examples as new augmented data, which also enhanced the ability of the model to resist adversarial attacks. However, the two stages of feature extraction and cross-modal fusion in existing methods are too independent. To deal with this problem, CMSA-MTPT \cite{CMSA-MTPT} proposed a multi-task pre-training framework to enhance the compatibility and applicability of the pre-trained features for cross-modal fusion, while also generating corresponding pseudo labels for unlabeled data to expand the dataset. Furthermore, VQAMix \cite{SAN-MEVF+VQAMix-C(BAN-MEVF+VQAMix-C)} proposed a simple yet effective data augmentation method that generated new samples by linearly combining any pair of Q\&A samples, while designing a conditional-mixed strategy that utilized language-type prior to force mixed samples to be the same category. Different from the pre-training task specifically designed by \cite{CMSA-MTPT} for Med-VQA, the pre-training tasks introduced in our MPMA framework are more universal, which not only achieve state-of-the-art performance in Med-VQA task (multi-modal) but also in medical image classification (uni-modal) and medical report generation tasks (cross-modal). The multi-task learning framework introduced in \cite{ma2021multitaskvqa} exhibits certain parallels with the framework we propose. However, in the context of intra-modality, \cite{ma2021multitaskvqa} only employs self-attention mechanisms from the Transformer architecture to capture region-level and word-level correlations. In contrast, our method not only captures the self-attention inherent relationships within ViT-based and BERT-based backbones but also incorporates intra-modality reconstruction to extract single-modal representation invariance.

\begin{figure*}[ht] 
	\centering 
	\includegraphics[width=\linewidth]{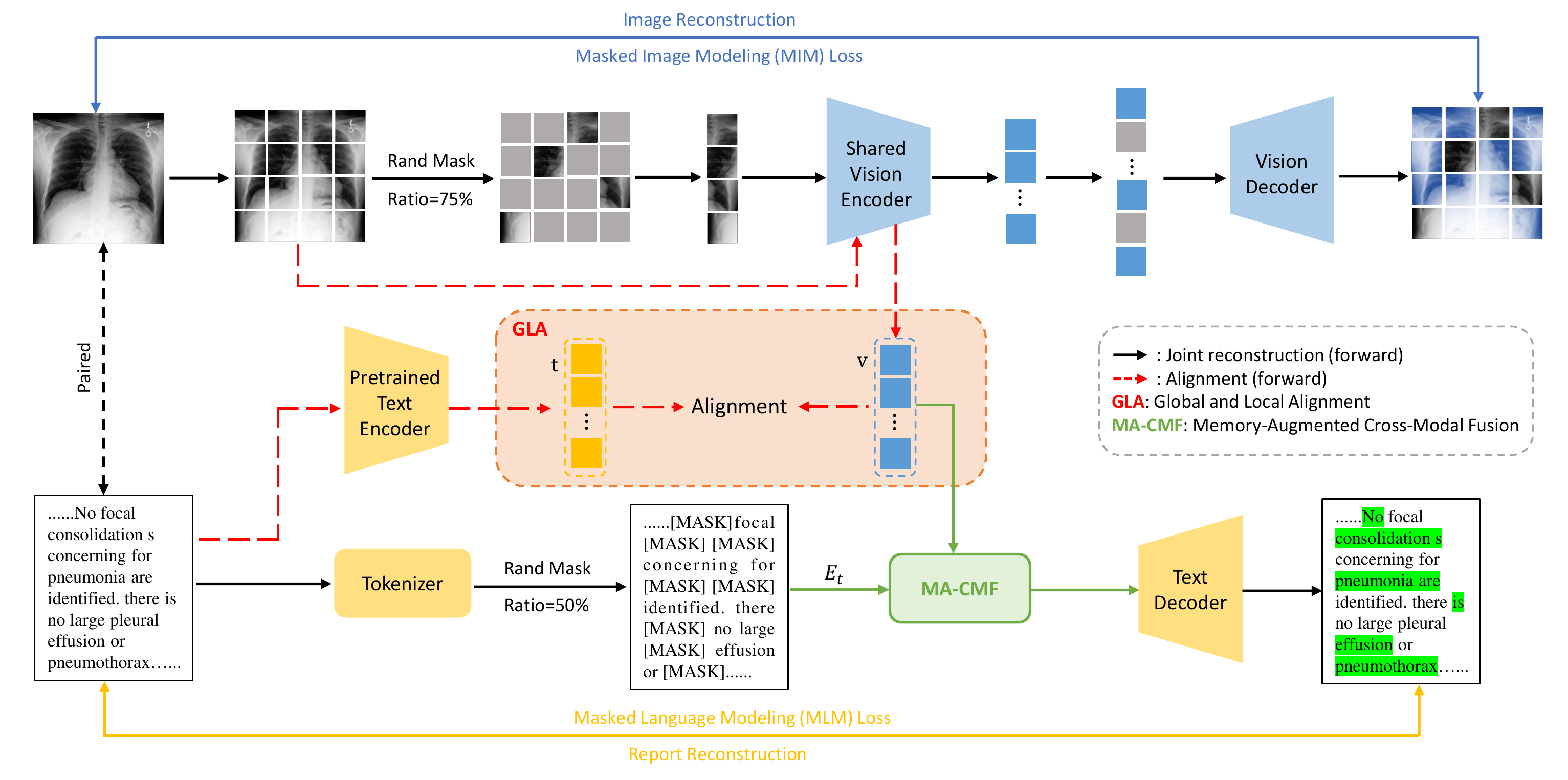} 
	\caption{Overview of our proposed MPMA framework for Med-VLP which consists of three pre-training tasks, namely image reconstruction task, report reconstruction task, and Global and Local Alignment (GLA) task. Specifically, GLA aligns the global visual feature $v$ and global report representation $t$. Additionally, the Memory-Augmented Cross-Modal Fusion (MA-CMF) module aggregates the global visual features $v$ with report embedding $E_t$ to obtain cross-modal representation for report reconstruction. Notably, the Share Vision Encoder takes the unmasked image patches as inputs for image reconstruction while taking the global image patches as inputs for Global and Local Alignment, separately.} 
	\label{fig:framework} 
\end{figure*}

\section{Method}
Fig. \ref{fig:framework} shows the overview of our proposed Multi-task Paired Masking with Alignment (MPMA) framework for Med-VLP, which consists of three pre-training tasks: image reconstruction task, report reconstruction task, and Global and Local Alignment (GLA) task. Specifically, the image reconstruction task follows MAE \cite{MAE} to mask image patches, which reduces the redundancy of visual information by masking patches in a large proportion; The report reconstruction task applies WordPiece \cite{wordpiece} as the tokenizer, and we propose a Memory-Augmented Cross-Modal Fusion (MA-CMF) module to fuse visual and report features to obtain cross-modal representations. With the assistance of visual information, cross-modal representations are fed into the text decoder for report reconstruction. Compared to BERT \cite{Bert}, we increase the masking ratio of report words to enable better cross-modal fusion and interaction with visual information. GLA task aligns paired images with reports using similarity calculation at both global and local scales and assists in the image and report reconstruction through adequate cross-modal interaction. In the reconstruction of images and reports, we leverage masked image modeling loss and masked language modeling loss to facilitate the process. Specifically, the image reconstruction task and the GLA task utilize the same vision encoder, while the report reconstruction task and the GLA task employ different text encoders. Our pre-training tasks are elaborated as follows. The main notations and definitions are shown in Table \ref{tab:notations} for better readability in the Appendix \ref{Notations}.

\subsection{Image Reconstruction}
Following MAE \cite{MAE}, we obtain the image representation by masking a large ratio of the image (i.e., 75\%) and then reconstructing it. The motivation is to cover more image regions to reduce the redundancy between them. In the image reconstruction branch, we use a pure vision transformer as the image encoder, denoting the input image as $I\in \mathbb{R}^{C\times H\times W}$, where $C$ is the number of channels, $H$ and $W$ are the height and width of the image, respectively. After the reshape operation, the image $I$ is split into multiple patches of the same size as $P\times P$ to obtain $ I\in \mathbb{R}^{N_p\times P^2C}$, where $N_p=\frac{H\times W}{P^2}$ is the number of patches. At this point, the image $I$ can be viewed as $N_p$ patch sequences of size $P^2\times C$. Then the patches are linearly projected into patch embeddings and add the corresponding position embedding $ E_{pos_1}\in \mathbb{R}^{N_p\times D}$, resulting in $E_v={IW_{l_1}}^T+E_{pos_1}$, where $W_{l_1}\in \mathbb{R}^{D\times P^2C}$ is a linear transformation. $h$ patches are selected to be randomly masked using random sampling, and the whole image $I$ can be grouped as masked tokens $I_m=\{{I}_m^1,I_m^2, \ldots ,I_m^h\}$ (ground truth) and visible unmasked tokens $I_u=\{{I}_u^1,I_u^2, \ldots ,I_u^{N_p-h}\}$, then only $I_u$ is fed into the shared vision encoder (denoted as $E_I$) to obtain its corresponding feature representation $I_U=E_I\left(I_u\right)=\{I_U^1,I_U^2, \ldots ,I_U^{N_p-h}\}$. After image encoding, in order to provide the original position information of the masked tokens, we fill the $I_U$ with position embedding to recover the original $N_p$ size and input it to the vision decoder (denoted as $D_I$) for image reconstruction. The reconstructed image patches are compared with the ground truth patches by the $\operatorname{MSE}$ loss function as follows:
\begin{equation}
    \mathcal{L}_{\mathrm{MIM}}\left(I_m,I_u\right)=\mathrm{MSE}\left(I_m^{1:h},D_I\left(E_I\left(I_u^{1:\left(N_p-h\right)}\right)\right)\right).
\end{equation}
For the sake of brevity, the process of filling position embedding with $E_I\left(I_u^{1:\left(N_p-h\right)}\right)$ is omitted from the equation.

\begin{figure}[t] 
	\centering 
	\includegraphics[width=\linewidth]{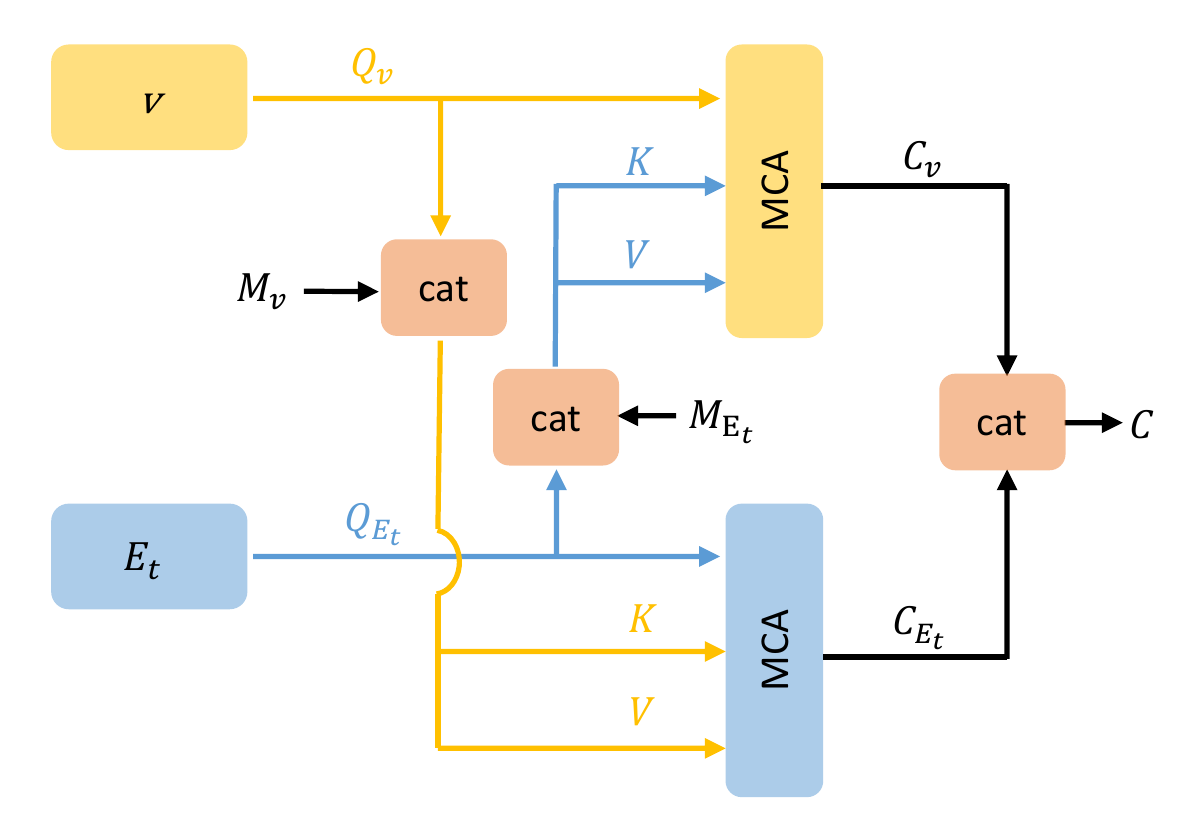} 
	\caption{The proposed Memory-Augmented Cross-Modal Fusion (MA-CMF) module aggregates the global visual features $v=E_I(I)$ with report embedding $E_t$ to obtain cross-modal representation $C$, where cat denotes the concatenation operation and MCA represents for Multi-head Cross-Attention.} 
	\label{fig:fusion} 
\end{figure}

\subsection{Report Reconstruction}
In the report reconstruction branch, reports and images appear in pairs, and we use WordPiece \cite{wordpiece} as the tokenizer to convert the report $R$ into tokens as $R=\{r^1,r^2, \ldots ,r^M\}\in \mathbb{R}^{M\times K}$, where the word token $r^w\in R^K$ is represented in one-hot format, and $K$ is the dimension of each word piece.
To reduce redundancy in textual information and better fuse cross-modal visual representations for report reconstruction, we mask a large ratio of 50\% of $R$ to obtain the masked part $R_m=\{r_m^1,r_m^2, \ldots ,r_m^n\}$ (ground truth) and the unmasked part $R_u=\{r_u^1,r_u^2,\ldots,r_u^{M-n}\}$. Compared to the 15\% masking ratio in BERT, this larger percentage of masking is expected to result in more informative representations.
For the unmasked report token $R_u$, we add $\left[MASK \right]$ at the corresponding position of the mask to keep the same size as the original report $R$ and project $R_u$ into a textual embedding $E_t={R_uW}_{l_2}^T+E_{pos_2}\in \mathbb{R}^{ M \times D}$, where the learnable parameters $W_{l_2}$ are randomly initialized with a fixed dictionary size, and $E_{pos_2}$ is the position embedding for the whole report. To complement textual information with visual information for report reconstruction, we propose a memory-augmented cross-modal fusion module $F(\cdot)$ (as shown in the Fig. \ref{fig:fusion}) to aggregate the global visual features $v=E_I(I)\in \mathbb{R}^{N_p\times D}$ with report embedding $E_t$ to obtain cross-modal representation $C$. Rather than directly concatenating the visual and textual embeddings, we adopt a progressive fusion approach to fully integrate cross-modal representations. Firstly, Multi-head Cross-Attention (MCA) is applied to the shallow pre-fusion between visual and textual modality features. MCA is a variant of Multi-head Self Attention (MSA) that is tailored to different input scenarios involving multiple modalities, and it can be formally expressed as follows:
\begin{equation}
   \operatorname{MCA}(Q, K, V)=\underbrace{\left(\operatorname{Softmax}\left(\frac{Q W_{i}^{Q} \cdot K W_{i}^{K}}{\sqrt{d}}\right)  V  W_{i}^{V}\right)   }_{ \operatorname{\underset{i\in\{1,...,h\}}{Concat} }} W^{M},
\end{equation}
where $W_i^Q,W_i^K,W_i^V\in \mathbb{R}^{d\times\frac{d}{h}}$ are the head projection matrices, $h$ is the number of parallel heads, and $W^M\in \mathbb{R}^{d\times d}$ represents the multi-head linear projection.
The global visual feature $v=E_I(I)\in \mathbb{R}^{N_p\times D}$ with textual embedding $E_t$ is fed into MCA to obtain the cross-modal intermediate representation $C_v$ and $C_{E_t}$. Meanwhile, to better record the modality patterns, we extend the key and value by adding learnable memory matrices, denoted as $M_v$ and $M_{E_t}$. Thus, the visual and textual pre-fusion processes are formulated as follows:
\begin{equation}
    C_v=\operatorname{MCA}\left(v,\left[E_t,M_{E_t}\right],\left[E_t,M_{E_t}\right]\right),
\end{equation}
\begin{equation}
    C_{E_t}=\operatorname{MCA}\left(E_t,\left[v,M_v\right],\left[v,M_v\right]\right),
\end{equation}
where $\left[,\right]$ is the concatenate operation. \par
The cross-modal global features $C=\left[C_v,C_{E_t}\right]$ can be obtained by concatenating the cross-modal intermediate representations $C_v$ and $C_{E_t}$, which are then fed into the text decoder (simple Transformer architecture) for report reconstruction. Finally, the training is carried out through the loss function of masked language modeling:
\begin{equation}
    \begin{aligned}
        \mathcal{L}_{\mathrm{MLM}}\left(R_{m}, E_{t}, I_{U}\right)=-\sum_{i=1}^{n}\log P & \Big(r_{m}^{i}  \mid E_{t}^{1: M}, I_{U}^{1:N_{p}} ;   \\ &  \ \  \Theta_{E_{I}}, \Theta_{D_{T}}, \Theta_{F} \Big),
    \end{aligned}
\end{equation}
where $\Theta_{E_{I}}, \Theta_{D_{T}}$ and $\Theta_{F}$ denote the weight parameters of the shared vision encoder, text decoder, and MA-CMF module, respectively.

\begin{figure}[t] 
	\centering 
	\includegraphics[width=\linewidth]{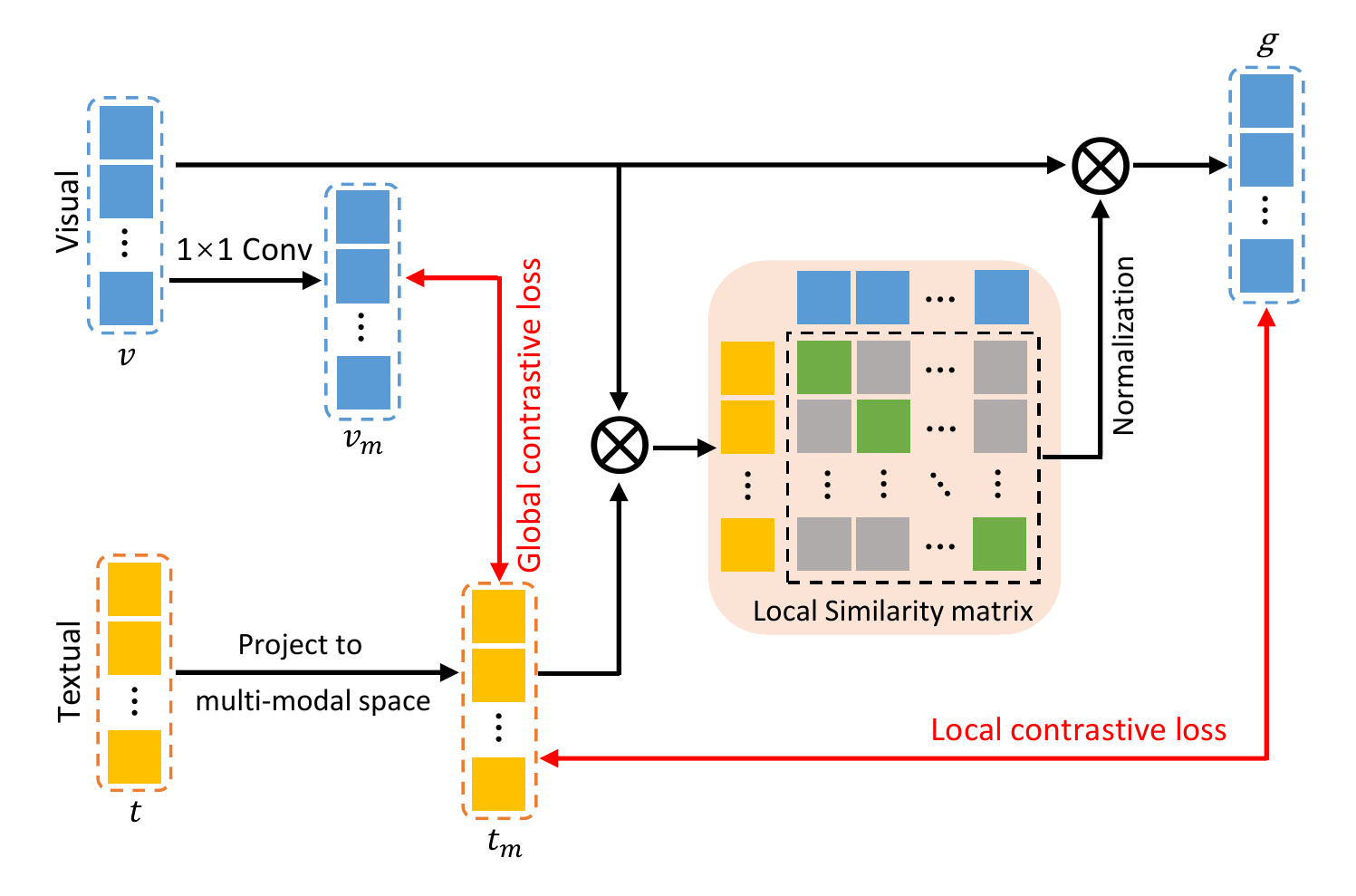} 
	\caption{The calculation process for Global and Local Alignment (GLA) module which models the matching of global and local representation.} 
	\label{fig:GLA} 
\end{figure}

\subsection{Global and Local Contrastive Alignment}
To effectively capture the inter-similarity and intra-similarity between paired medical images and reports, we propose a novel global and local alignment module, which models the matching of both global and local representations. As illustrated in Fig. \ref{fig:GLA}, the global and local alignment module shares the same vision encoder with the image reconstruction task. However, unlike the image reconstruction task, the input image is directly fed into the shared vision encoder without random masking after patchifying. This process produces a global visual representation $v\in \mathbb{R}^{N_p \times D}$ that corresponds to the image. Meanwhile, the report paired with the image is input into a pre-trained text encoder to obtain global report representations $t\in \mathbb{R}^{M \times K}$. As shown in Fig. \ref{fig:GLA}, to compare global representations in cross-modal space, we first use $1\times 1$ convolution to project $v$ into cross-modal space to obtain a cross-modal representation of global visual features $v_m=W_v\cdot v\in \mathbb{R}^{M\times D}$, where $W_v\in \mathbb{R}^{M\times N_p}$. Then, the global report features are projected into the $D$-dimensional cross-modal space to obtain $t_m=t\cdot W_t^T \in \mathbb{R}^{M\times D}$, where $W_t\in \mathbb{R}^{D\times K}$. Denoting $\left(v_i^m,t_i^m\right), i\in \{ 1,2,\ldots,N \}$ as one of paired images and reports among $N$ pairs.
Our goal is to align the global representations between paired $v_m$ and $t_m$ as closely as possible. To achieve this goal, we formulate a symmetric global contrastive loss $\mathcal{L}_{\mathrm{g}}$ that minimizes the negative log posterior probability to align the inter-similarity between the paired images and reports:
\begin{equation}
    \begin{aligned}
         \mathcal{L}_{\mathrm{g}}=-\frac{1}{N}\ \sum_{i=1}^{N} & \left( \log\ \frac{\exp\ \left(\mathbf{v}_{i}^{\mathrm{m}}\ \cdot\ \mathbf{t}_{i}^{\mathrm{m}}\ /\ \tau_{1}\right)}{\sum_{j=1}^{N}\ \exp\ \left(\mathbf{t}_{i}^{\mathrm{m}}\ \cdot\ \mathbf{v}_{j}^{\mathrm{m}}\ /\ \tau_{1}\right)}  \right.\\  &\left.  + \log\ \frac{\exp\ \left(\mathbf{v}_{i}^{\mathrm{m}}\ \cdot\ \mathbf{t}_{i}^{\mathrm{m}}\ /\ \tau_{1}\right)}{\sum_{j=1}^{N}\ \exp\ \left(\mathbf{v}_{i}^{\mathrm{m}}\ \cdot\ \mathbf{t}_{j}^{\mathrm{m}}\ /\ \tau_{1}\right)}\right),
    \end{aligned}
\end{equation}
where $\tau_1$ is a scaling factor to control the temperature, and $N$ is the batch size. \par

For local representation learning, in order to measure the similarity between image regions and report words, we apply dot products to construct a similarity matrix $s=v\cdot t_m^T\in \mathbb{R}^{N_p\times M}$. The element $s_{ij}$ in this matrix measures the degree of correlation between image region $v_i$ and the report word $t_j$. To eliminate the influence of the similarity scale, we apply regularization to obtain a local normalized similarity coefficient matrix $\alpha_{ij}$ as follows:
\begin{equation}
    \alpha_{ij}=\frac{\exp{s_{ij}}/\tau_2}{\sum_{k=1}^{N_p}\exp{s_{kj}}/\tau_2},
\end{equation}
where $\tau_2$ is a scaling factor to control the temperature. \par
Similarity coefficient $\alpha_{ij}$ is utilized to calculate the visual aggregation feature related to the $j$-th word $g_j={\sum_{i=1}^{N_p}\alpha}_{ij}v_i$. 
Our goal is to align the visual aggregation feature $g_j$ and the multi-modal report word feature $t_j$ as closely as possible. To achieve this, we calculate the local representation similarity $H(\cdot)$ using the following formula:
\begin{equation}
    H\left(X_g,X_t\right)=\log{\sum_{j=1}^{M}\exp{\left(g_j\cdot\ t_j\right)}},
\end{equation}
where $g_j$ and $t_j$ correspond to the $j$-th element in $X_g$ and $X_t$, respectively. \par
We extend the calculation of local representation similarity to a batch N and design local contrastive loss $\mathcal{L}_{\mathrm{l}}$ for learning local representation alignment and aligning the intra-similarity between paired images and reports:
\begin{equation}
    \begin{aligned}
         \mathcal{L}_{\mathrm{l}}=-\frac{1}{N} \sum_{i=1}^{N} &\left(\log \frac{\exp \left(\mathbf{H}(\mathbf{X_g}_i,\mathbf{X_t}_i)/ \tau_{3}\right)}{\sum_{j=1}^{N} \exp \left(\mathbf{H}(\mathbf{X_g}_j,\mathbf{X_t}_i)/ \tau_{3}\right)}  \right.\\  &\left.  +\log \frac{\exp \left(\mathbf{H}(\mathbf{X_g}_i,\mathbf{X_t}_i)/ \tau_{3}\right)}{\sum_{j=1}^{N} \exp \left(\mathbf{H}(\mathbf{X_g}_i,\mathbf{X_t}_j) / \tau_{3}\right)}\right),
    \end{aligned}
\end{equation}
where $\tau_3$ is a scaling factor to control the temperature, and $N$ is the batch size. \par

\subsection{Multi-task Training Scheme}
We adopt multi-task learning to integrate the loss functions of the pre-training tasks for training. The overall objective function $\mathcal{L}_{\mathrm{all}}$ consists of four loss terms, namely the image reconstruction loss $\mathcal{L}_{\mathrm{MIM}}$, the report reconstruction loss $\mathcal{L}_{\mathrm{MLM}}$, and the global and local contrastive losses $\mathcal{L}_{\mathrm{g}}$ and $\mathcal{L}_{\mathrm{l}}$:
\begin{equation}
    \mathcal{L}_{\mathrm{all}}= \mathcal{L}_{\mathrm{MIM}}+\lambda_{IL} \mathcal{L}_{\mathrm{MLM}}+\lambda_{GLA} \left( \mathcal{L}_{\mathrm{g}}+\lambda_{gl} \mathcal{L}_{\mathrm{l}} \right) ,
\end{equation}
where the hyper-parameter $\lambda_{IL}$ balances the loss terms of the image reconstruction and report reconstruction tasks, while $\lambda_{gl}$ balances the global and local contrastive loss terms. The hyper-parameter $\lambda_{GLA}$ determines the weight between the GLA task and the joint image-text reconstruction task. Their values will be further discussed in the experiment.

\section{Experiments and Results}
To verify the effectiveness of our proposed MPMA framework for medical vision-language pre-training, we fine-tune and conduct extensive experiments on three medical downstream tasks, i.e., medical image classification, medical report generation, and medical VQA tasks. The comparisons with the state-of-the-art methods show the effectiveness and generalizability of our MPMA in multiple medical tasks.

\subsection{Data for Pre-training}
\label{Data for Pre-training}
We conducted model pre-training on the MIMIC-CXR \cite{mimic} and ROCO \cite{ROCO} datasets, the descriptions are as follows:

\begin{itemize}[leftmargin=*]
	\item \textbf{MIMIC-CXR:} In the experiment, we use the official splits to partition the dataset into a training set with 368960 images and 222758 reports, a validation set with 2991 images and 1808 reports, and a testing set with 5159 images and 3269 reports. We focus on paired images and reports, so we only retained the frontal radiographs images with corresponding reports for model pre-training on the training set.
	\item \textbf{ROCO:} It includes image-caption pairs collected from PubMed Central which is an open-access biomedical literature database. It contains over 81000 non-compound medical images with corresponding captions which are randomly split into 65,460/8,183/8,182 for training/validation/testing set. We filter non-radiology samples.
\end{itemize}

\subsection{Evaluation Tasks \& Fine-tuning Data}
\subsubsection{Medical Image Classification}

\begin{itemize}[leftmargin=*]
        \item \textbf{CheXpert} \cite{chexpert}: A multi-label classification task for Chest X-rays has been defined. We adopt the official setting \cite{chexpert} and selected the most representative 5 lesions for reporting, ie., atelectasis, cardiomegaly, consolidation, edema, and pleural effusion. Because the official testing set of CheXpert is not open-access, we follow ConVIRT \cite{ConVIRT} to use the official validation set as the testing set and randomly sampled 5000 images from the training set as the validation set. The training/validation/testing set is divided into 218414/5000/234 images, respectively.
	\item \textbf{RSNA Pneumonia} \cite{RSNAPneumonia}: A binary classification task has been defined for Chest X-rays, the dataset contains 30k frontal view Chest radiographs labeled as either pneumonia or normal. We use the official splits to divide the training/validation/test set into 25184/1500/3000 images.
 \end{itemize}
 
\subsubsection{Medical Report Generation}

\begin{itemize}[leftmargin=*]
        \item \textbf{IU X-Ray} \cite{IU}: Following previous works \cite{HRGR,KERP}, we exclude the images without reports and randomly split them by the ratio of 7:2:1 into training, validation, and testing set.
	\item \textbf{MIMIC-CXR} \cite{mimic}: We only use the training set divided in Section \ref{Data for Pre-training} to fine-tune the model. And the performance is tested on the corresponding testing set.
 \end{itemize}
 
\subsubsection{Medical VQA}

\begin{itemize}[leftmargin=*]
        \item \textbf{VQA-RAD} \cite{VQA-RAD}: It contains 3515 question-answer (QA) pairs generated by clinicians, which is split into 3064 QA pairs as the training set and 451 QA pairs as the test set. There are 11 categories of clinical questions: abnormality, attribute, color, count, modality, organ, plane, positional reasoning, object/condition presence, size, and others. Questions can be divided into closed-ended questions with limited choices and open-ended questions without a limited structure.
	\item \textbf{Path-VQA} \cite{PATH-VQA}: It contains 4,998 pathology images with 32,795 question-answer pairs, which are collected from PEIR digital library and the pathology book. It includes eight types of questions, which are closed-form, ‘what’, ‘where’, ‘when’, ‘how’, ‘why’, ‘whose’, and ‘how much’. The number of questions corresponding to an image varies from 1 to 10.
 \end{itemize}

\begin{table*}[ht]
\centering
\setlength\tabcolsep{10pt}
\caption{Comparative experiments of medical image classification on CheXpert and RSNA Pneumonia datasets. We report AUC scores under different labeling ratios when fine-tuning on CheXpert and RSNA Pneumonia. * denotes GLoRIA is implemented by ViT-B/16.}
\begin{tabular}{lcclll|lll}
\toprule
\multirow{2}{*}{Method} & \multirow{2}{*}{\begin{tabular}[c]{@{}c@{}}Input\\ Size\end{tabular}} & \multirow{2}{*}{\begin{tabular}[c]{@{}c@{}}Pre-trained\\ Data\end{tabular}} & \multicolumn{3}{c}{CheXpert}   & \multicolumn{3}{c}{RSNA Pneumonia} \\   \cmidrule(r){4-6}  \cmidrule(l){7-9}
                        &                                                                       &                                                                             & 1\%      & 10\%     & 100\%    & 1\%        & 10\%      & 100\%     \\   \midrule
ConVIRT \cite{ConVIRT}                & 224                                                                   & CheXpert                                                                    & 85.9     & 86.8     & 87.3     & 77.4       & 80.1      & 81.3      \\
GLoRIA \cite{gloria}                 & 224                                                                   & CheXpert                                                                    & 86.6     & 87.8     & 88.1     & 86.1       & 88.0      & 88.6      \\
ConVIRT \cite{ConVIRT}                & 224                                                                   & MIMIC-CXR                                                                   & 87.0     & 88.1     & 88.1     & 88.8       & 91.5      & 92.7      \\
MedKLIP \cite{medklip}                & 224                                                                   & MIMIC-CXR                                                                   & -        & -        & -        & 87.3       & 88.0      & 89.3      \\ \midrule
BioViL \cite{BioviL}                 & 480                                                                   & \begin{tabular}[c]{@{}c@{}}PubMed+\\ MIMIC-CXR\end{tabular}                & -        & -        & -        & 88.1       & 88.4      & 89.1      \\ \midrule
GLoRIA* \cite{gloria}                & 224                                                                   & MIMIC-CXR                                                                   & 86.5\tiny$\pm$0.8 & 87.5\tiny$\pm$0.6 & 87.8\tiny$\pm$0.5 & 89.7\tiny$\pm$0.8   & 91.2\tiny$\pm$0.5  & 92.1\tiny$\pm$0.3  \\
REFERS \cite{REFERS}                 & 224                                                                   & MIMIC-CXR                                                                   & 87.2\tiny$\pm$0.8 & 88.1\tiny$\pm$0.5 & 88.2\tiny$\pm$0.3 & 89.4\tiny$\pm$0.7   & 91.6\tiny$\pm$0.7  & 92.7\tiny$\pm$0.4  \\
M3AE \cite{M3AE}                   & 224                                                                   & MIMIC-CXR                                                                   & 86.2\tiny$\pm$0.6 & 87.3\tiny$\pm$0.6 & 87.9\tiny$\pm$0.4 & 89.0\tiny$\pm$0.5   & 90.8\tiny$\pm$0.6  & 92.3\tiny$\pm$0.3  \\
MGCA \cite{MGCA}                   & 224                                                                   & MIMIC-CXR                                                                   & 88.8     & 89.1     & 89.7     & 89.1       & 89.9      & 90.8      \\
MRM \cite{MRM}                    & 224                                                                   & MIMIC-CXR                                                                   & 88.5\tiny$\pm$0.7 & 88.5\tiny$\pm$0.6 & 88.7\tiny$\pm$0.3 & \textbf{91.3}\tiny$\pm$0.6   & 92.7\tiny$\pm$0.4  & 93.3\tiny$\pm$0.4  \\ \midrule
MPMA (ours)                     & 224                                                                   & MIMIC-CXR                                                                   & \textbf{89.1}\tiny$\pm$0.8 & \textbf{89.8}\tiny$\pm$0.5 & \textbf{90.6}\tiny$\pm$0.4 & \textbf{91.3}\tiny$\pm$0.6   & \textbf{93.4}\tiny$\pm$0.5  & \textbf{94.1}\tiny$\pm$0.3  \\
\bottomrule
\end{tabular}
\label{tab:classification comparison}
\end{table*}

\begin{table*}[ht]
\centering
\setlength\tabcolsep{1.5pt}
\caption{Comparative experiments of medical report generation on IU X-Ray and MIMIC-CXR datasets. The results of NLG metrics are reported when fine-tuning on IU X-Ray and MIMIC-CXR. B-n, M, and R-L denote BLEU-n, METEOR, and ROUGE-L, respectively. * denotes pre-trained models implemented by official code finetuning with Transformer as the report decoder.}
\begin{tabular}{lcccccc|cccccc}
\toprule              
\multirow{2}{*}{Method} & \multicolumn{6}{c}{IU   X-Ray}                                                    & \multicolumn{6}{c}{MIMIC-CXR}                                                     \\   \cmidrule(r){2-7}  \cmidrule(l){8-13}
                        & B-1         & B-2         & B-3         & B-4         & M           & R-L         & B-1         & B-2         & B-3         & B-4         & M           & R-L         \\  \midrule
Show-Tell \cite{show-tell}              & 0.346       & 0.214       & 0.141       & 0.095       & -           & 0.320       & 0.299       & 0.184       & 0.121       & 0.084       & -           & 0.263       \\
Att2in \cite{Att2in}                 & 0.399       & 0.249       & 0.172       & 0.126       & -           & 0.321       & 0.325       & 0.203       & 0.136       & 0.096       & -           & 0.276       \\
AdaAtt \cite{AdaAtt}                 & 0.436       & 0.288       & 0.203       & 0.150       & -           & 0.354       & 0.299       & 0.185       & 0.124       & 0.088       & -           & 0.266       \\
Transformer \cite{transformer}            & 0.422       & 0.264       & 0.177       & 0.120       & -           & 0.338       & 0.314       & 0.192       & 0.127       & 0.090       & -           & 0.265       \\
R2Gen \cite{R2Gen}                  & 0.470       & 0.304       & 0.219       & 0.165       & 0.187       & 0.371       & 0.353       & 0.218       & 0.145       & 0.103       & 0.142       & 0.277       \\
R2GenCMN \cite{R2GenCMN}               & 0.475       & 0.309       & 0.222       & 0.170       & 0.191       & 0.375       & 0.353       & 0.218       & 0.148       & 0.106       & 0.142       & 0.278       \\
PPKED \cite{PPKED}                  & 0.483       & 0.315       & 0.224       & 0.168       & -           & 0.376       & 0.360       & 0.224       & 0.149       & 0.106       & 0.149       & 0.284       \\
AlignTrans \cite{aligntransformer}             & 0.484       & 0.313       & 0.225       & 0.173       & 0.204       & 0.379       & 0.378       & 0.235       & 0.156       & 0.112       & 0.158       & 0.283       \\
CMM+RL \cite{CMM+RL}                 & 0.494       & 0.321       & 0.235       & \textbf{0.181}       & 0.201       & 0.384       & 0.381       & 0.232       & 0.155       & 0.109       & 0.151       & 0.287       \\
RAMT-U \cite{RAMT-U}                 & 0.482       & 0.310       & 0.221       & 0.165       & 0.195       & 0.377       & 0.362       & 0.229       & 0.157       & 0.113       & 0.153       & 0.284       \\
ConVIRT* \cite{ConVIRT}                 & 0.473       & 0.308       & 0.220       & 0.169       & 0.185       & 0.368       & 0.355       & 0.217       & 0.143       & 0.101       & 0.140       & 0.275       \\
MGCA* \cite{MGCA}                 & 0.485       & 0.325       & 0.230       & 0.172       & 0.193       & 0.371       & 0.376       & 0.224       & 0.152       & 0.108       & 0.149       & 0.282       \\
Clinical-BERT \cite{clinical-bert}           & 0.495       & 0.330       & 0.231       & 0.170       & -           & 0.376       & 0.383       & 0.230       & 0.151       & 0.106       & 0.144       & 0.275       \\
Multi-Criteria \cite{Multi-Criteria}         & 0.496       & 0.319       & 0.241       & 0.175       & -           & 0.377       & 0.351       & 0.223       & 0.157       & 0.118       & -           & 0.287       \\ \midrule
MPMA (ours)                     & \textbf{0.518}\tiny$\pm$0.005 & \textbf{0.337}\tiny$\pm$0.005 & \textbf{0.253}\tiny$\pm$0.003 & 0.179\tiny$\pm$0.001 & \textbf{0.220}\tiny$\pm$0.003 & \textbf{0.388}\tiny$\pm$0.006 & \textbf{0.392}\tiny$\pm$0.004 & \textbf{0.246}\tiny$\pm$0.004 & \textbf{0.166}\tiny$\pm$0.002 & \textbf{0.122}\tiny$\pm$0.003 & \textbf{0.164}\tiny$\pm$0.003 & \textbf{0.295}\tiny$\pm$0.002   \\
\bottomrule
\end{tabular}
\label{tab:report generation comparison}
\end{table*}

\subsection{Experimental Settings}
For the implementation, we adopt CLIP-ViT-B \cite{CLIP-ViT-B} as the shared vision encoder and CXR-BERT \cite{BioviL} as the pre-trained text encoder. We use center-crop to resize all images to 224$\times$224, and the patch size of ViT is 16$\times$16 with a total of 14$\times$14 patches for each image. The masking ratio is set to 75\% for images and 50\% for reports. More experimental details can be found in Appendix \ref{More Implementation Details}. \par
For the evaluation metrics, we follow the previous studies to adopt AUC scores for medical image classification, accuracy for medical VQA, and NLG metrics (i.e., BLEU \cite{bleu}, METEOR \cite{meteor} and ROUGE-L \cite{rouge}) for medical report generation.

\begin{table*}[ht]
\centering
\setlength\tabcolsep{10pt}
\caption{Comparative experiments of medical VQA on VQA-RAD and PATH-VQA datasets. Mean accuracy and standard deviation are reported when fine-tuning on VQA-RAD and PATH-VQA. $\dag$ \ denotes methods using data augmentations. $\ast$ denotes our implementation of pre-trained models using their official code, and fusion is operated at the end of the pre-trained models with the addition of Transformer as the decoder.}
\begin{tabular}{llll|lll}
\toprule
\multirow{2}{*}{Method} & \multicolumn{3}{c}{VQA-RAD (\%)}    & \multicolumn{3}{c}{PATH-VQA (\%)}   \\  \cmidrule(r){2-4} \cmidrule(l){5-7}
                        & Open     & Closed   & Overall  & Open     & Closed   & Overall  \\   \midrule
SAN \cite{SAN}                    & 24.2     & 57.2     & 44.0     & 1.6      & 59.4     & 30.5     \\
BAN \cite{BAN}                    & 27.6     & 66.5     & 51.0     & 2.9      & 68.2     & 35.6     \\
SAN-MAML \cite{SAN-MAML(BAN-MAML)}               & 38.2     & 69.7     & 57.1     & 5.4      & 75.3     & 40.5     \\
BAN-MAML \cite{SAN-MAML(BAN-MAML)}               & 40.1     & 72.4     & 60.7     & 5.9      & 79.5     & 42.9     \\
SAN-MEVF \cite{SAN-MEVF(BAN-MEVF)}               & 40.7     & 74.1     & 60.8     & 6.0      & 81.0     & 43.6     \\
BAN-MEVF \cite{SAN-MEVF(BAN-MEVF)}               & 43.9     & 75.1     & 62.6     & 8.1      & 81.4     & 44.8     \\
BAN-MEVF+CR \cite{BAN-MEVF+CR}            & 52.4\tiny$\pm$0.9 & 79.3\tiny$\pm$1.1 & 68.5\tiny$\pm$1.0 & -        & -        & -        \\
CMSA-MTPT \cite{CMSA-MTPT}              & 52.8\tiny$\pm$1.8 & 77.8\tiny$\pm$0.4 & 67.9\tiny$\pm$0.8 & -        & -        & -        \\
SAN-MEVF+MMQ \cite{SAN-MEVF+MMQ(BAN-MEVF+MMQ)}           & 46.3\tiny$\pm$1.8 & 73.0\tiny$\pm$1.4 & 62.3\tiny$\pm$1.1 & 9.6\tiny$\pm$0.5  & 83.7\tiny$\pm$0.4 & 46.8\tiny$\pm$0.3 \\
BAN-MEVF+MMQ \cite{SAN-MEVF+MMQ(BAN-MEVF+MMQ)}           & 52.0\tiny$\pm$1.1 & 72.4\tiny$\pm$0.9 & 64.3\tiny$\pm$0.7 & 11.8\tiny$\pm$0.6 & 82.1\tiny$\pm$0.5 & 47.1\tiny$\pm$0.4 \\
BAN-MEVF+DAVQA\dag \cite{BAN-MEVF+DAVQA}         & 51.2\tiny$\pm$1.4 & 76.2\tiny$\pm$1.4 & 66.2\tiny$\pm$1.3 & 8.5\tiny$\pm$0.4  & 83.4\tiny$\pm$0.2 & 46.1\tiny$\pm$0.2 \\
BAN-MEVF+SEADA\dag \cite{BAN-MEVF+SEADA}         & 49.6\tiny$\pm$2.0 & 72.4\tiny$\pm$1.5 & 63.3\tiny$\pm$1.2 & 9.1\tiny$\pm$0.5  & 81.3\tiny$\pm$0.3 & 45.3\tiny$\pm$0.4 \\
MTPT \cite{MTPT}               & 61.5     & 80.9     & 73.2     & -      & -     & -     \\
MMBERT \cite{mmbert}               & 63.1     & 77.9     & 72.0     & -      & -     & -     \\
ConVIRT* \cite{ConVIRT}      & 62.2\tiny$\pm$1.3 & 75.9\tiny$\pm$1.0 & 70.8\tiny$\pm$1.7 & 10.2\tiny$\pm$0.6 & 82.7\tiny$\pm$0.5 & 46.9\tiny$\pm$0.3 \\
MGCA* \cite{MGCA}      & 64.8\tiny$\pm$1.6 & 80.5\tiny$\pm$1.2 & 73.6\tiny$\pm$1.3 & 14.0\tiny$\pm$0.8 & 85.1\tiny$\pm$0.4 & 49.2\tiny$\pm$0.5 \\
SAN-MEVF+VQAMix-C\dag \cite{SAN-MEVF+VQAMix-C(BAN-MEVF+VQAMix-C)}      & 53.8\tiny$\pm$1.9 & 74.0\tiny$\pm$2.4 & 65.9\tiny$\pm$1.9 & 12.1\tiny$\pm$0.5 & 84.4\tiny$\pm$0.2 & 48.4\tiny$\pm$0.2 \\
BAN-MEVF+VQAMix-C\dag \cite{SAN-MEVF+VQAMix-C(BAN-MEVF+VQAMix-C)}      & 56.6\tiny$\pm$1.3 & 79.6\tiny$\pm$1.5 & 70.4\tiny$\pm$1.1 & 13.4\tiny$\pm$0.6 & 83.5\tiny$\pm$0.2 & 48.6\tiny$\pm$0.3 \\
M3AE* \cite{M3AE}                  & 65.4\tiny$\pm$1.4 & 81.7\tiny$\pm$1.2 & 74.2\tiny$\pm$0.9 & 14.2\tiny$\pm$0.7 & 84.0\tiny$\pm$0.5 & 48.8\tiny$\pm$0.4 \\ \midrule
MPMA (ours)                     & \textbf{68.2}\tiny$\pm$1.1 & \textbf{84.2}\tiny$\pm$1.4 & \textbf{77.6}\tiny$\pm$0.8 & \textbf{16.4}\tiny$\pm$0.5 & \textbf{86.8}\tiny$\pm$0.5 & \textbf{50.2}\tiny$\pm$0.3  \\
\bottomrule
\end{tabular}
\label{tab:medical VQA comparison}
\end{table*}

\begin{figure*}[ht] 
	\centering 
	\includegraphics[width=\linewidth]{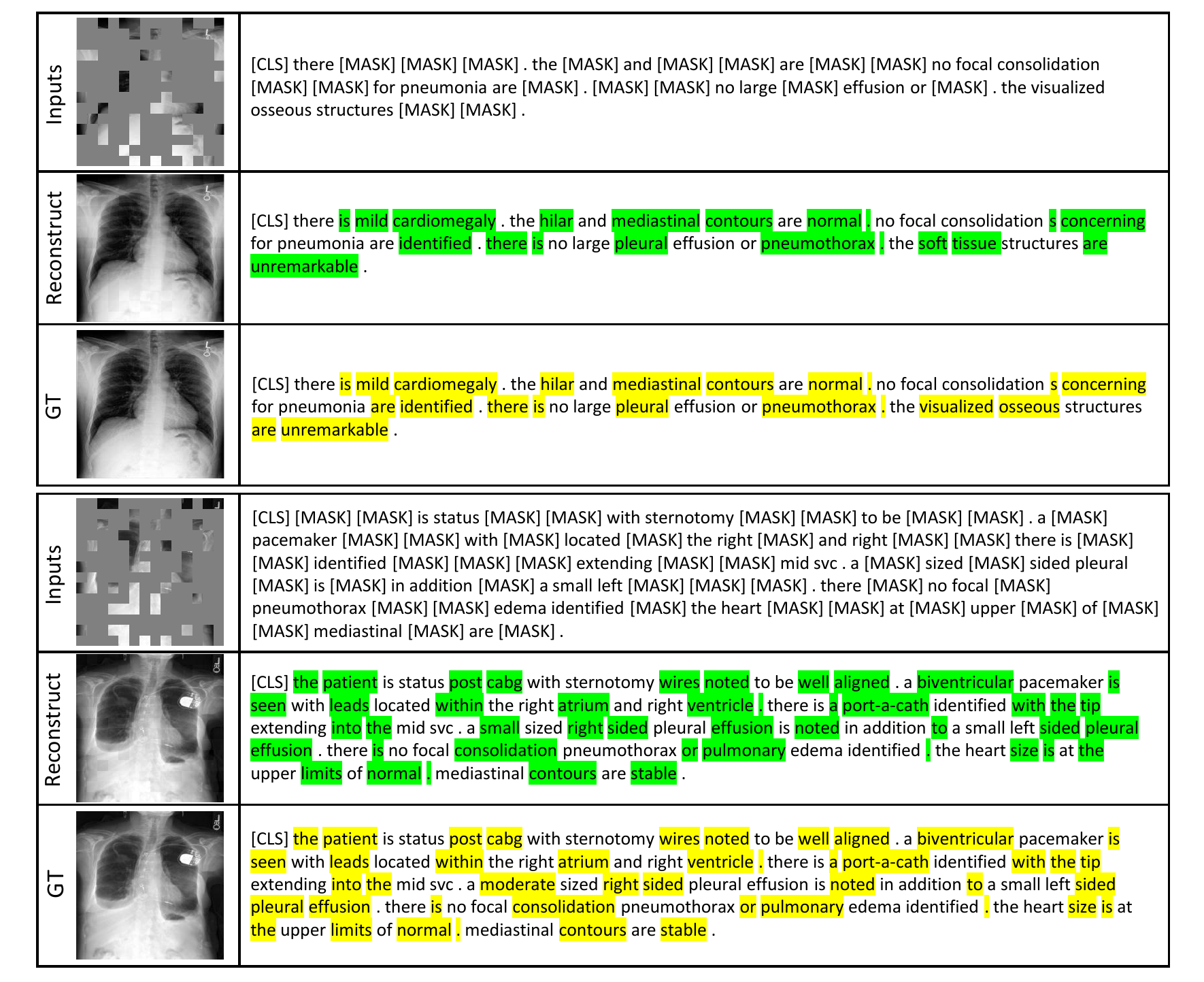} 
	\caption{Paired masking and reconstruction examples on MIMIC-CXR dataset. For each case, we present the masked image with reports (Inputs), our reconstructed pairs (Reconstruct), and the ground truth (GT). Reconstructed and corresponding words are highlighted in green and yellow, respectively.} 
	\label{fig:reconstruct case} 
\end{figure*}

\begin{figure*}[ht] 
	\centering 
	\includegraphics[width=\linewidth]{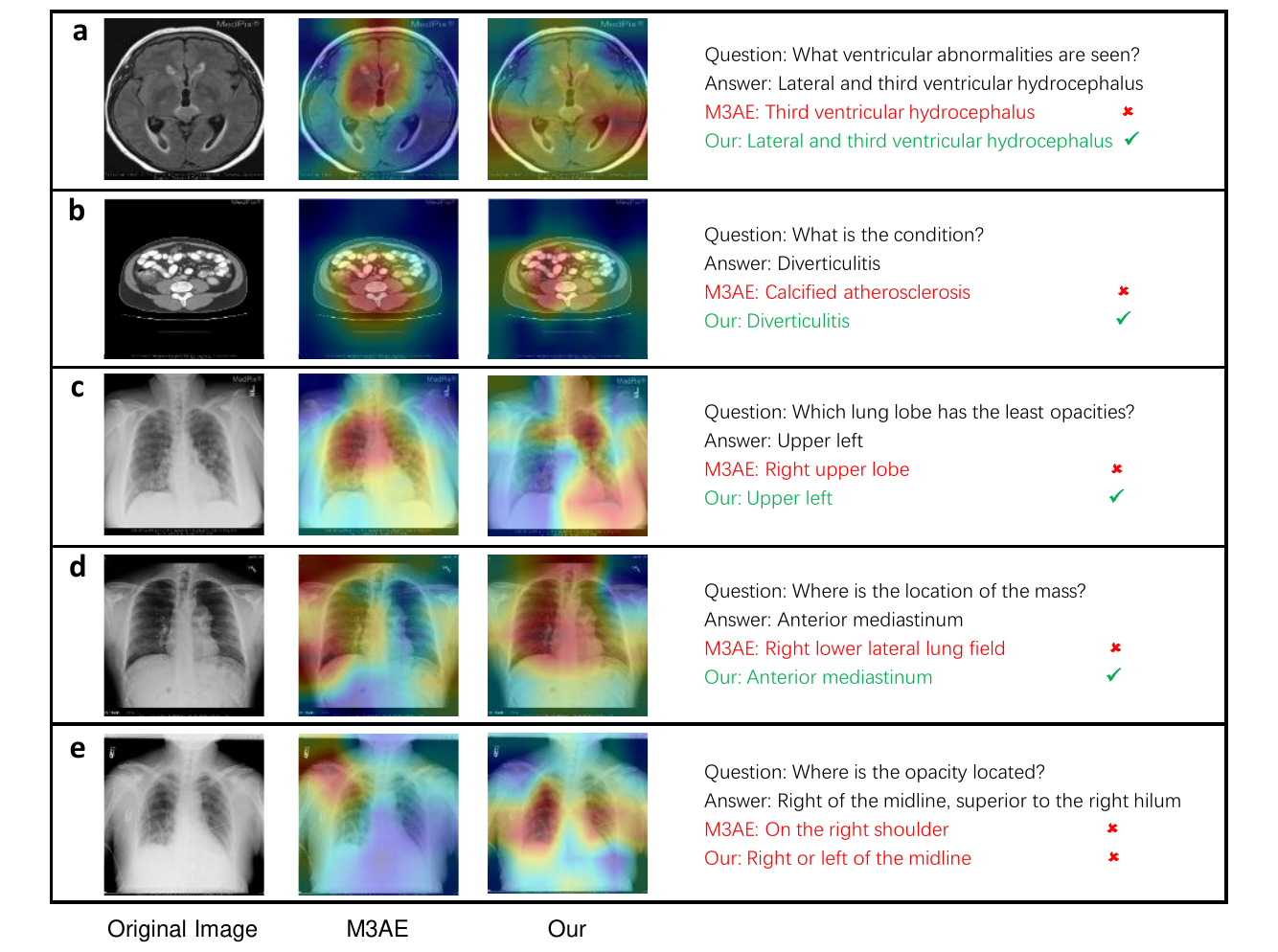} 
	\caption{Several visualized results on the Med-VQA task between the M3AE and our proposed framework. These examples are selected from CT, MRI, and X-Ray modalities of the VQA-RAD dataset. Red and Green denote the wrong and correct predictions, respectively.} 
	\label{fig:VQA case} 
\end{figure*}

\subsection{Comparison With State-of-the-Art Methods}
\subsubsection{Medical Image Classification}
We fine-tuned our proposed MPMA framework using different ratios of labeled data for medical image classification tasks on the CheXpert and RSNA Pneumonia datasets. We report classification AUC scores and compare the performance with several advanced state-of-the-art pre-training approaches, including ConVIRT \cite{ConVIRT}, GloRIA \cite{gloria}, MedKLIP \cite{medklip}, BioViL \cite{BioviL}, REFERS \cite{REFERS}, M3AE \cite{M3AE}, MGCA \cite{MGCA}, and MRM \cite{MRM}. Comparison results are quoted from the original papers or \cite{MRM}. As shown in Table \ref{tab:classification comparison}, MPMA achieves significant performance improvement compared to SOTA Med-VLP methods under various label ratios on both datasets. Specifically, when the label ratios are quite small, i.e., 1\% and 10\%, our MPMA outperforms M3AE and REFERS by 2.9\%, 2.5\% and 1.9\%, 1.7\% on the CheXpert dataset, and by 2.3\%, 2.6\% and 1.9\%, 1.8\% on the RSNA Pneumonia dataset, respectively. This indicates that our pre-training method can still demonstrate strong transferability even when fine-tuning with a small amount of labeled data. Moreover, when fine-tuning with the full ratio of labeled data, our MPMA still outperforms M3AE by 1.7\% and 2.8\% on the two datasets, respectively. This demonstrates the effectiveness of our pre-training strategy that integrates alignment and reconstruction tasks. Notably, our MPMA also outperforms the advanced MRM method on both datasets, which also models reconstruction tasks, highlighting the effectiveness of our proposed multi-task modeling scheme.
\subsubsection{Medical Report Generation}
We fine-tuned our proposed MPMA on medical report generation datasets, namely IU X-Ray and MIMIC-CXR, and evaluated the generated reports using the widely-used evaluation toolkit \cite{toolkit}. We calculated the matching degree between the generated reports and ground truth reports using NLG metrics, including BLEU \cite{bleu}, METEOR \cite{meteor}, and ROUGE-L \cite{rouge}. We compared our proposed model with current state-of-the-art methods for medical report generation, including pre-training methods such as Clinical-BERT \cite{clinical-bert} and non-pre-training SOTA methods such as R2Gen \cite{R2Gen}, R2GenCMN \cite{R2GenCMN}, PPKED \cite{PPKED}, AlignTrans \cite{aligntransformer}, CMM+RL \cite{CMM+RL}, RAMT-U \cite{RAMT-U}, and Multi-Criteria \cite{Multi-Criteria}. The comparison results are quoted from the original papers. As shown in Table \ref{tab:report generation comparison}, our MPMA achieves the best or second-best performance compared to the state-of-the-art methods on both datasets. Specifically, compared to the pre-trained Clinical-BERT, our MPMA has improved by 5.3\% and 3.2\% on BLUE-4 and ROUGE-L on the IU X-Ray dataset. On the MIMIC-CXR dataset, our MPMA has improved by 15.1\% and 13.9\% on BLUE-4 and METEOR, respectively. These improvements fully demonstrate the effectiveness of integrating alignment and reconstruction tasks in the pre-training stage. Compared to the latest non-pre-training method Multi-Criteria, our MPMA significantly outperforms in several metrics, especially in BLEU-1, BLEU-4, and ROUGE-L, with improvements of 11.7\%, 3.4\%, and 2.9\% on the MIMIC-CXR dataset, respectively. This indicates that the multi-task pre-training scheme has indeed helped the vision encoder to obtain better feature representation, resulting in more accurate and informative generated reports. It is worth noting that our MPMA achieves a greater improvement on BLUE-1 compared to other metrics. The possible reason is that the local alignment in our proposed GLA module enables better alignment between local single words and local single regions, leading to a higher single-word hit rate.
\subsubsection{Medical VQA}
We fine-tuned our MPMA on VQA-RAD and PATH-VQA datasets and compared its performance with state-of-the-art medical VQA methods, including pre-training approaches such as M3AE \cite{M3AE}, MMBERT \cite{mmbert}, and MTPT \cite{MTPT}, as well as non-pre-training methods such as VQAMix \cite{SAN-MEVF+VQAMix-C(BAN-MEVF+VQAMix-C)}, BAN-MEVF+SEADA \cite{BAN-MEVF+SEADA}, BAN-MEVF+DAVQA \cite{BAN-MEVF+DAVQA} and CMSA-MTPT \cite{CMSA-MTPT}, etc. Among them, VQAMix, BAN-MEVF+SEADA, and BAN-MEVF+DAVQA apply data augmentations to improve performance. Comparison results are quoted from the original papers or \cite{SAN-MEVF+VQAMix-C(BAN-MEVF+VQAMix-C)}. We report mean accuracy on closed-ended and open-ended questions with overall accuracy in Table \ref{tab:medical VQA comparison}, which shows that our proposed MPMA achieves the best overall accuracy on both VQA-RAD and PATH-VQA datasets. Specifically, compared to the latest advanced pre-training model M3AE, our MPMA showed an improvement by 2.8\%, 2.5\%, and 3.4\% in closed-ended, open-ended questions and overall accuracy on VQA-RAD dataset, and an improvement by 2.2\%, 2.8\% and 1.4\% on PATH-VQA dataset, respectively. These improvements fully demonstrate the effectiveness of integrating alignment and reconstruction tasks in pre-training. Compared to the complex data augmentation method VQAMix without pre-training, our MPMA significantly outperforms it by 11.6\%, 4.6\%, and 7.2\% in closed-ended, open-ended questions and overall accuracy on VQA-RAD dataset, and 3.0\%, 3.3\% and 1.6\% improvements on PATH-VQA dataset, respectively. This indicates that our method not only outperforms the current advanced pre-training methods but also outperforms methods specifically designed for medical VQA tasks. Overall, pre-training methods outperform non-pre-training methods in various metrics, benefiting from the learning process of generic feature mapping. Meanwhile, it is worth noting that our MPMA shows a particularly significant improvement in accuracy compared to VQAMix on open-ended questions. A possible explanation is that our proposed multi-task aggregation paradigm has a better-promoting effect on open-ended questions compared to VQAMix, which relies solely on data augmentation.

\subsection{Reconstruction Analysis}
To provide a more intuitive understanding of the reconstruction task in the pre-training stage, we visualize the masking and reconstruction process of both the image and report modalities. Fig. \ref{fig:reconstruct case} presents paired masking and reconstruction examples on the MIMIC-CXR dataset. It can be observed that even with high masking ratios, our MPMA framework can still restore the vast majority of image and report content. However, in some cases, some details may be lost. For example, in the second example, the text reconstruction incorrectly describes the degree of pleural effusion as "small-sized" instead of "moderate-sized". The reason being that the bottom area of the chest cavity in the radiograph is tightly covered, which indicates that the reconstruction of the report requires the assistance of radiograph content, highlighting the importance of introducing the proposed cross-modal fusion module. In the first example, it is interesting to note that “visualized osseous” was incorrectly identified as “soft tissue”. A possible explanation for this error is that “soft tissue” appears more frequently in the sentence structure of "... are unremarkable" than "visualized osseous". Additionally, the sternal structures in the image are comprehensively obscured, which may have contributed to the error. Additional experiments on the exploration of how global image feature complement report reconstruction are presented in Appendix \ref{Additional Analysis}.

\subsection{Qualitative Analysis}
To provide a qualitative comparison of the medical VQA task between our proposed MPMA and M3AE \cite{M3AE}, we use Grad-CAM \cite{gradcam} for interpretability and present visualized examples from three modalities, i.e., MRI, CT, and X-Ray, on different organs or tissues. As shown in Fig. \ref{fig:VQA case}, the first column is the original images, the second column presents the visualization results of M3AE, the third column displays the visualization results of our MPMA, and the last column is the corresponding Q\&A pairs. \par
In the first row (a), a brain MRI image is presented, and the questions are related to ventricular abnormalities. M3AE only focuses on the third venturar hydrocephalus, while our MPMA not only focuses on third venturar hydrocephalus but also identifies abnormalities in the lateral ventricular hydrocephalus. The second row (b) presents an abdomen CT image where M3AE locates the area around the spine, resulting in an incorrect answer of "calibrated spherosclerosis". In contrast, our MPMA correctly answers the question by identifying the ascending colon as the affected area. In the third row (c), a chest X-Ray image is presented, and our MPMA accurately identifies the position of the lung lobe at the upper left, while M3AE provides an answer that is completely opposite. The reason for this discrepancy is that M3AE only locates the right upper lobe as shown in the visualization results. In the fourth row (d), a chest X-Ray image is used to question the position of the mass, where M3AE locates the incorrect area of the right lower lateral lung field, while our MPMA provides a more accurate position through a broader receptive field. In the last row (e) of the presented examples, a chest X-ray image is presented with a question about the location of the opacity. Unfortunately, neither M3AE nor our MPMA could provide a completely correct answer. Compared to M3AE positioning the opacity on the right shoulder, our MPMA provides a relatively closer answer by focusing on the right and left areas of the midline, which overlaps with the correct answer. The reason for the inaccurate location of “superior to the right hilum” may be that the opacity of the image is relatively dense which can be induced by lung inflammation or tumors, and our MPMA still has errors in determining this highly fine-grained position. These examples demonstrate that our MPMA can more accurately focus on the reasonable regions of the image to answer questions with the help of cross-modal reconstruction and alignment. Still, there is a long way to go to achieve better interpretability.


\begin{table}[t]
\centering
\setlength\tabcolsep{3pt}
\caption{Ablation study on the four key components (i.e., image reconstruction loss ($\mathcal{L}_{\mathrm{MIM}}$), report reconstruction loss ($\mathcal{L}_{\mathrm{MLM}}$), global and local alignment module (GLA) and memory-augmented cross-modal fusion module (MA-CMF)) on CheXpert (AUC reported), MIMIC-CXR (R-L reported) and VQA-RAD (overall accuracy reported) datasets.}
\begin{tabular}{cccc|ccc}
\toprule
$\mathcal{L}_{\mathrm{MIM}}$   & $\mathcal{L}_{\mathrm{MLM}}$   & GLA     & MA-CMF  & CheXpert & MIMIC-CXR & VQA-RAD \\ \cmidrule(lr){1-4} \cmidrule(lr){5-7}
$\surd$ &         &         &         & 84.3     & 0.260     & 70.8    \\
$\surd$ & $\surd$ & $\surd$ &         & 87.5     & 0.279     & 74.3    \\
$\surd$ & $\surd$ &         & $\surd$ & 88.7     & 0.286     & 75.0    \\
$\surd$ & $\surd$ & $\surd$ & $\surd$ & \textbf{90.6}     & \textbf{0.295}     & \textbf{77.6}    \\
\bottomrule
\end{tabular}
\label{tab: ablation of components}
\end{table}

\begin{figure}[t] 
	\centering 
	\includegraphics[width=\linewidth]{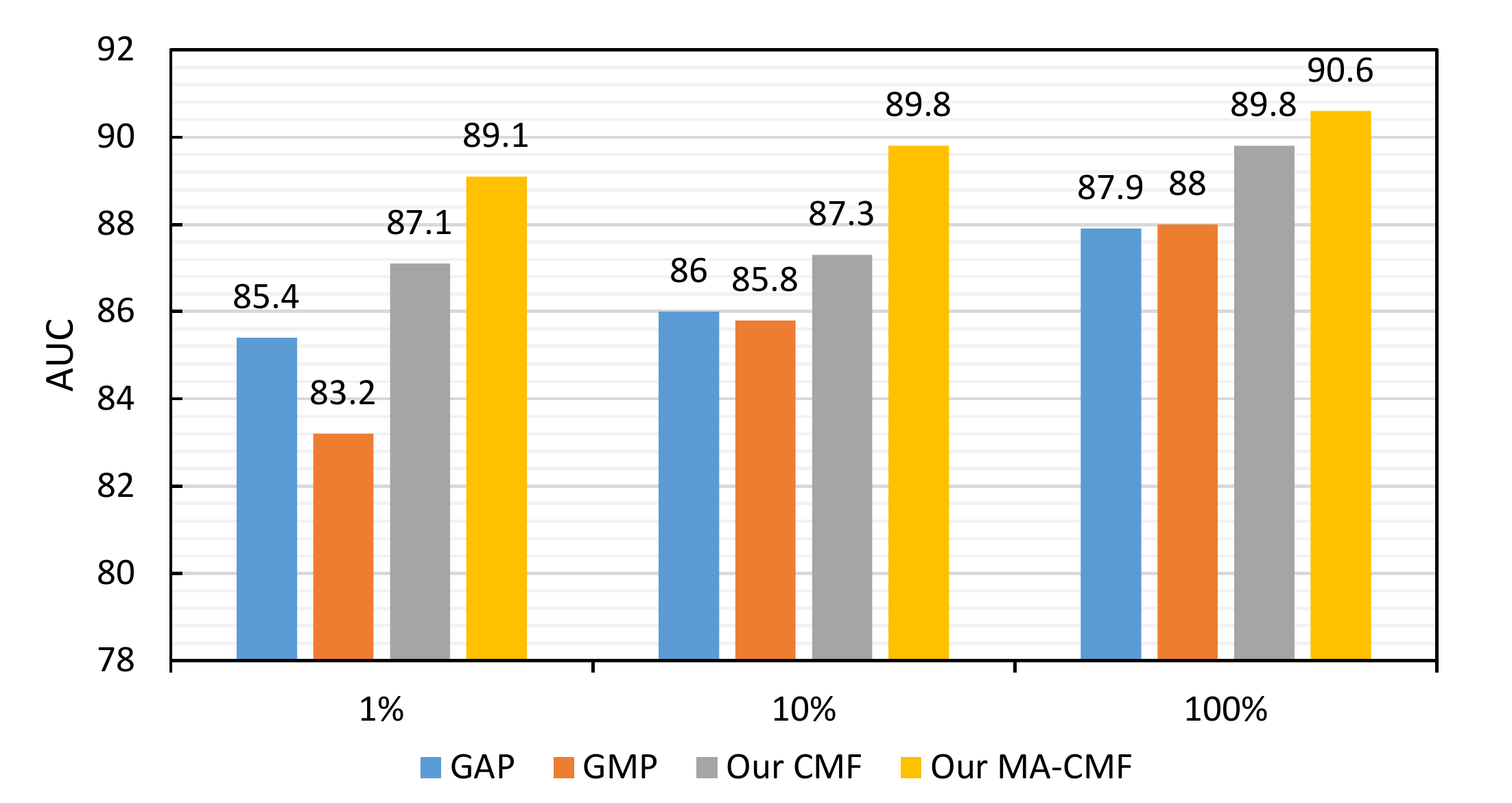} 
	\caption{Ablation study on the fusion strategy when integrating visual and textual representations for report reconstruction on the CheXpert dataset under various labeling ratios. AUC scores are reported.} 
	\label{fig:ablation of fusion strategy} 
\end{figure}

\subsection{Ablation Study}
To evaluate the contribution of each component in our proposed framework MPMA, we conduct an ablation study on the four key components, including image reconstruction loss ($\mathcal{L}_{\mathrm{MIM}}$), report reconstruction loss ($\mathcal{L}_{\mathrm{MLM}}$), global and local alignment module (GLA), and memory-augmented cross-modal fusion module (MA-CMF). As shown in Table \ref{tab: ablation of components}, we conduct experiments on all three downstream tasks, reporting AUC scores on the CheXpert dataset, R-L Metrics on the MIMIC-CXR dataset, and overall accuracy on the VQA-RAD dataset. To establish a baseline for comparison, we retain only image reconstruction $\mathcal{L}_{\mathrm{MIM}}$. However, the performance of the baseline sharply decreases across all three datasets, highlighting the importance of the other three components. When adding report construction $\mathcal{L}_{\mathrm{MLM}}$ and MA-CMF to the baseline, we can observe significant improvements of 5.2\%, 10\%, and 5.9\% on classification, report generation, and VQA tasks, respectively. This suggests that pairwise cross-modal fusion after report reconstruction significantly facilitates the image reconstruction task. Notably, the report generation task on MIMIC-CXR achieves a larger boost compared to other tasks, indicating that more adequate cross-modal interaction leads to better downstream task adaptation, particularly for the image-to-text modality report generation task. Furthermore, the addition of report reconstruction $\mathcal{L}_{\mathrm{MLM}}$ and GLA module to the baseline also results in significant performance improvements compared to the baseline on all three tasks, with an AUC score increase from 84.3 to 87.5, R-L increasing from 0.260 to 0.279, and overall accuracy increasing from 70.8 to 74.3. This demonstrates the effectiveness of cross-modal alignment for pre-training. Moreover, the integration of GLA and MA-CMF, which constitutes our whole MPMA framework, further increases performance. This suggests that GLA and MA-CMF can mutually reinforce each other, and adequate cross-modal alignment can better assist in reconstructing the semantic invariance of the task modeling.

Additionally, we further perform an ablation study to analyze the influence of applying different cross-modal fusion strategies, including global average pooling (GAP), global maximum pooling (GMP), our MA-CMF and its version with the memory part removed (our CMF). As shown in Fig. \ref{fig:ablation of fusion strategy}, our CMF and MA-CMF outperform GAP and GMP by a significant margin, demonstrating the effectiveness of our proposed cross-modal fusion module. Both GAP and GMP show significant performance degradation under only 1\% labeling ratio, particularly GMP. However, our proposed fusion strategy maintains relatively stable performance even with very few labeled data.
Furthermore, it can be observed that MA-CMF shows some performance improvement compared to CMF, indicating that the addition of memory does help better record the modality pattern and assist in accomplishing a more adequate cross-modal fusion.

\begin{table}[t]
\centering
\setlength\tabcolsep{10pt}
\caption{Sensitivity analysis of hyper-parameters on CheXpert (AUC reported), MIMIC-CXR (R-L reported), and VQA-RAD (overall accuracy reported) datasets. $\lambda_{IL}$ and $\lambda_{gl}$ are varied for the investigation. The default values are $\lambda_{IL}=5$ and $\lambda_{gl}=1$.}
\begin{tabular}{ccccc}
\toprule
$\lambda_{IL}$ & $\lambda_{gl}$ & CheXpert & MIMIC-CXR & VQA-RAD \\  \cmidrule(lr){1-2} \cmidrule(lr){3-5}
1              & \multicolumn{1}{c|}{1}              & 89.9     & 0.288     & 77.2    \\
5              & \multicolumn{1}{c|}{1}              & \textbf{90.6}     & \textbf{0.295}     & \textbf{77.6}    \\
10             & \multicolumn{1}{c|}{1}              & 88.5     & 0.289     & 76.2    \\
1              & \multicolumn{1}{c|}{5}              & 88.2     & 0.273     & 74.9    \\
\bottomrule
\end{tabular}
\label{tab: sensitivity of hyper-parameters}
\end{table}

\begin{figure}[t] 
	\centering 
	\includegraphics[width=\linewidth]{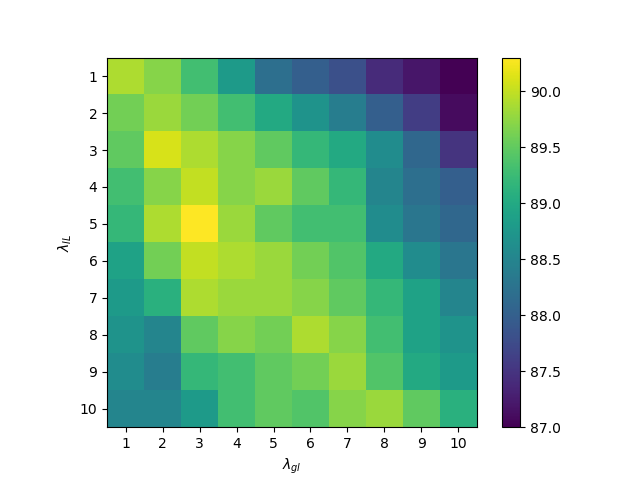} 
	\caption{AUC values for different $\lambda_{IL}$ and $\lambda_{gl}$ on CheXpert dataset.} 
	\label{fig:hyperparameter} 
\end{figure}

\subsection{Sensitivity Analysis}
\label{Sensitivity Analysis}
As shown in Table \ref{tab: sensitivity of hyper-parameters}, we investigate the sensitivity of different hyperparameter settings for $\lambda_{IL}$ and $\lambda_{gl}$ on three datasets with different tasks. It can be observed that the performance is not highly sensitive when $\lambda_{IL}$ is greater than or equal to $\lambda_{gl}$. However, setting $\lambda_{gl}$ to 5 while fixing the value of $\lambda_{IL}$ results in a significant drop in performance across all three datasets. This is likely due to the increased weight of the alignment task in multi-task training, which could be detrimental to the reconstruction task that may not have yet stabilized in the early stages of model training. This finding also supports our earlier conjecture. We observe that the optimal performance is achieved when $\lambda_{IL}$ is equal to 5 and $\lambda_{gl}$ is equal to 1. \par
To further investigate the impact of different values of $\lambda$, we conduct a grid search on $\lambda_{IL}$ and $\lambda_{gl}$ with values ranging from 1 to 10 on the CheXpert dataset. Our results, as shown in Fig. \ref{fig:hyperparameter}, indicate that the best performance is achieved when $\lambda_{IL}=5$ and $\lambda_{gl}=3$, indicating that both terms contribute to the improved results. Based on our ablation study and hyperparameter tuning experiments, we set $\lambda_{IL}$ and $\lambda_{gl}$ to 5 and 3, respectively, in our experiments.

\section{Conclusion}
This work is the first attempt to integrate the cross-modal alignment task into a joint medical image-text reconstruction framework to achieve more comprehensive cross-modal interaction. Our proposed global and local alignment (GLA) module is designed to assist the self-supervised paradigm in obtaining semantic representations with rich domain knowledge. Additionally, we have introduced a Memory-Augmented Cross-Modal Fusion (MA-CMF) module to achieve more comprehensive fusion of visual features and to assist in the process of report reconstruction. Extensive experiments on six widely-used public benchmark datasets across three downstream tasks demonstrate the superiority of our approach, even when using only a few labeled pairs. \par
Although our method performs well in downstream tasks, there are also some limitations. For example, there exists data bias in the dataset, with data categories exhibiting a severe long tail distribution, and CT and MRI modalities only appearing in some parts of the ROCO dataset, resulting in uneven distribution of data for each modality. Besides, we only use the simplest Transformer as the text decoder for textual generation, which may limit the text generation capability of our proposed framework.\par
For future work, to overcome above mentioned limitations, we suggest the introduction of resampling or active learning to balance the dataset, which is especially crucial in the medical domain to alleviate data bias issues. Additionally, it would be interesting to explore the use of currently popular pre-trained large language models (e.g., LLAMA \cite{LLAMA}, UL2 \cite{UL2}) to enhance the diversity of textual expressions and potentially achieve further performance improvements. These directions could further improve the effectiveness and efficiency of vision-language pre-training frameworks for medical tasks.



\bibliographystyle{IEEEtran}
\bibliography{./bibliography}

\clearpage
\onecolumn
\begin{appendices}
\setcounter{table}{0}   
\setcounter{figure}{0}
\renewcommand{\thetable}{C\Roman{table}}
\renewcommand{\thefigure}{B\Roman{figure}}


\section{More Implementation Details}
\label{More Implementation Details}
We first pre-train the proposed MPMA framework on MIMIC-CXR and ROCO datasets and then fine-tune it on six downstream datasets among three medical tasks to verify the quality of pre-training. 

\subsection*{Vision and Text Encoders}

For the image files, we center-crop and resize the images to have a size of $224\times224$. CLIP-ViT-B$\footnote{\url{https://huggingface.co/openai/clip-vit-base-patch16}}$ \cite{CLIP-ViT-B} as the shared vision encoder split the image with patch size $16\times16$, which results in 196 visual tokens for each image. Similar to ConVIRT \cite{ConVIRT}, We apply affine transformation with a degree sampled from [-20, 20], max horizontal and vertical translation fractions of 0.1, and a scaling factor sampled from [0.95, 1.05]; color jittering with brightness and contrast adjustment ratios sampled from [0.6, 1.4]; and Gaussian blur with $\sigma \in$ [0.1, 3.0]. Unlike ConVIRT  and GLoRIA \cite{gloria}, we do not apply horizontal flipping during joint training to preserve location information (e.g., “pneumothorax in the left upper lung”). Then, we prepend a learnable embedding ([CLS] token embedding) to the sequence of embedded patches and feed them into the vision transformer. The embedding of [CLS] token in the last layer represents image-level feature (768-d vector), and the embeddings of patches in the last layer represent visual token-level features $f_v\in \mathbb{R}^{768\times196}$. The embedding dimension of ViT-B is 768.

For the textual radiology report data, we first tokenize all reports with the CXR-BERT-specialized$\footnote{\url{https://huggingface.co/microsoft/BiomedVLP-CXR-BERT-specialized}}$ tokenizer which has been pre-trained on three stages. Next, we keep only the Findings and Impression sections and remove all other sections. We remove all image-text pairings from the dataset where the text section is empty or has less than 3 tokens. This preprocessing procedure gives us about 426k total image-text pairs for pretraining our framework. We truncate each report or pad it with [PAD] tokens to make all of them equal in size. The embedding of [CLS] token in the last layer represents the report-level feature (768-d vector), and embeddings of word tokens in the last layer represent text token-level features $f_t\in \mathbb{R}^{768\times128}$. 

\subsection*{More Training Details}

For pre-training, we use AdamW \cite{AdamW} rather than SGD as the optimizer with the learning rate of 2e-4 and weight decay set to 0.05. The batch size is set to 16. MSE loss and Cross-Entropy loss are used for masking image and language modeling, respectively. We pre-train for 30 epochs and early stop if total validation loss does not decrease after 5 straight runs, and the checkpoint with the lowest validation loss is saved for fine-tuning in the next step. The global and local alignment task with two contrasting losses will ramp up from 0 to 1 in the first T epochs to avoid prematurely forced alignment before the reconstruction task becomes stable. The coefficient $\lambda_{GLA}$ is determined by a Gaussian warming up function as $\lambda_{GLA\left(t\right)}=1\ast e^{-5\left(1-t/T\right)^2}$ and will be fixed to 1 after ramping up. $\lambda_{IL}$ and $\lambda_{gl}$ are set to 5 and 3 respectively, which has been further investigated in Section \ref{Sensitivity Analysis}. \par
For fine-tuning, AdamW is used as the optimizer with a learning rate of 5e-6 and 2e-4 for the previous model and the task-specific layers, respectively. The batch size is set to 16. For medical image classification (uni-modal) and medical report generation (cross-modal), we adopt pre-trained shared vision encoders as the vision encoders in the fine-tuning stage. Specifically, for medical image classification, we adopt a MLP layer after the shared vision encoder to perform multi-label classification; for medical report generation, a Transformer is added as the report decoder followed by the vision encoder. For medical VQA (multi-modal), we adopt the shared vision encoder and a pre-trained text encoder (CXR-BERT) as image and question encoders in the fine-tuning stage, respectively, while applying our proposed MA-CMF to integrate image and question representation. Then, the integrated features are fed into a Transformer for answering generation. We fine-tune the above models for 50 epochs and early stop when validation loss does not decrease for 10 straight runs, and save the checkpoint model with the lowest validation loss for test.\par

\section{Additional Analysis}
\label{Additional Analysis}
\begin{figure*}[t] 
	\centering 
	\includegraphics[width=\linewidth]{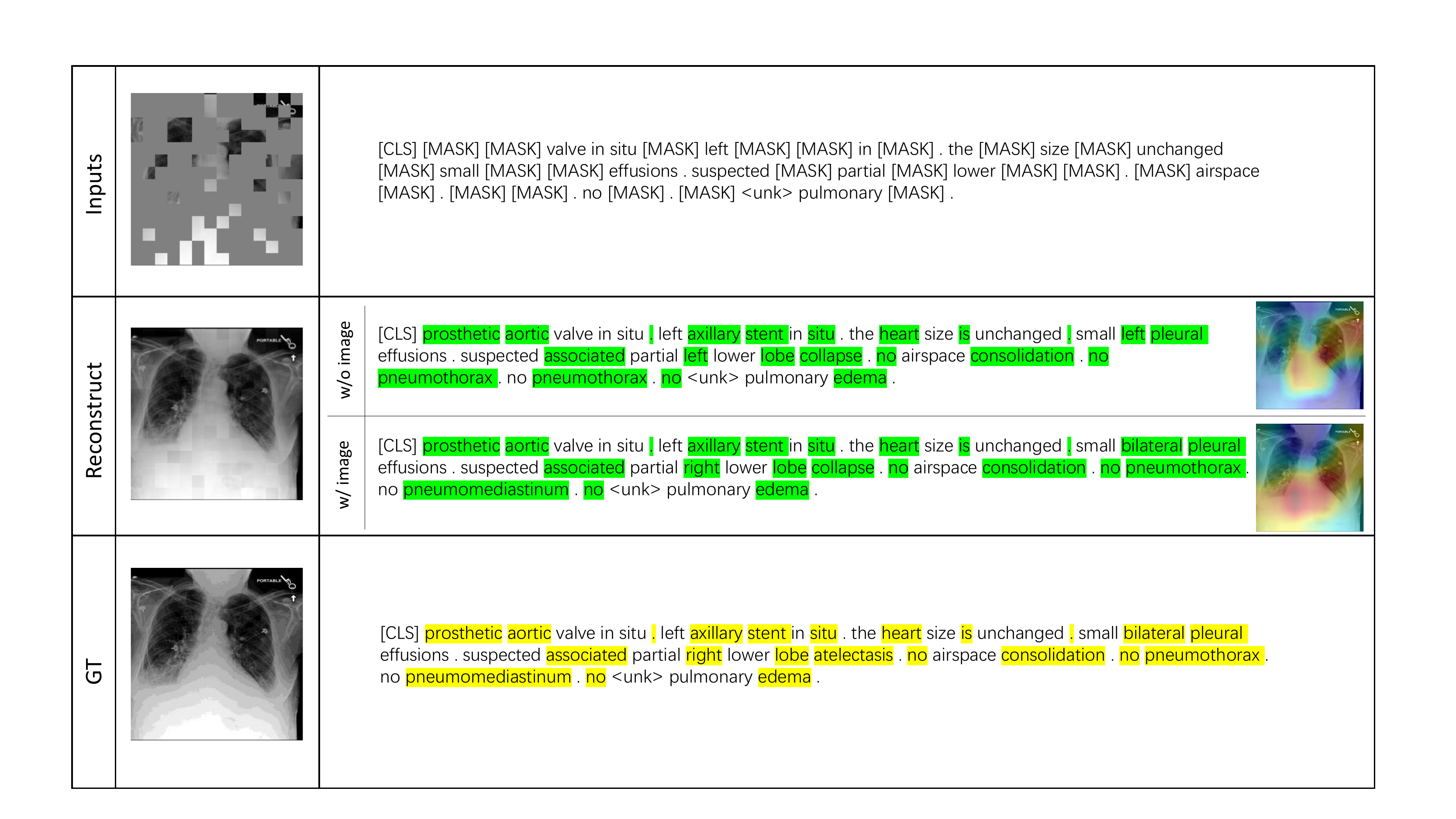} 
	\caption{Additional experiment on paired masking and reconstruction example on MIMIC-CXR dataset with visualization by heat maps. We present the masked image with reports (Inputs), our reconstructed pairs (Reconstruct), and the ground truth (GT). Reconstructed and corresponding words are highlighted in green and yellow, respectively. In the reconstruction part, with (w/ for short) image and without (w/o for short) image denotes whether the global image features are integrated into the report reconstruction process.} 
	\label{fig:additional-experiment} 
\end{figure*}

To better analyze how the global image feature complement report reconstruction in the reconstruction part, we conduct comparative experiments based on whether global image features are included in the report reconstruction. As shown in the middle row of Fig. \ref{fig:additional-experiment}, we perform report reconstruction separately with (w/ for short) and without (w/o for short) image fused, and present visualization of them.
By comparing the reconstruction reports in two different scenarios, it can be observed that when the global image feature is not added, the reconstructed report provides some inaccurate or even completely opposite position information. For example, the position of the pleural effusion should be bilateral rather than simply left, and the lower lobe should be located on the right rather than the left. Through the visualized heat maps, we can also clearly observe that the reconstructed focus of the w/o image scenario is completely concentrated on the left side of the lung. However, after the addition of image features, the focus area expands significantly to both sides, and the receptive field becomes more open. In addition, the w/o image provides a repetitive expression of “no pneumothorax”, possibly due to the lack of visual information assistance, MLM tasks tend to give words with higher frequency of occurrence, where “pneumothorax” appears significantly more frequently than “pneumothediastinum”. In these cases, adding global image features can fully supplement position-related information and to some extent avoid the repetition of high-frequency words, proving the effectiveness of integrating global image features in report reconstruction.

\section{Notations}
\label{Notations}

\begin{table*}[ht]
\centering
\renewcommand\arraystretch{1.1}
\setlength\tabcolsep{6pt}
\caption{Main Notations And The Definitions.}
\begin{tabular}{ll}
\toprule
\textbf{Notations}                                             & \textbf{Definitions}                                                                                         \\    \hline
$I$                                                   & the   input image                                                                                   \\
$N_p$                                                 & number   of image patches                                                                           \\
$D$                                                   & the   dimension of the image patch embedding                                                        \\
$I_m$                                                 & masked image tokens (a total of $h$)                                                                \\
$I_u$                                                 & unmasked image tokens (a total of   $N_p-h$)                                                        \\
$E_I$                                                 & the shared vision encoder in MPMA   framework                                                       \\
$I_U$                                                 & feature representation produced by   $E_I$ when feeding with  $I_u$                                 \\
$v$                                                   & global visual features produced by   $E_I$ when feeding with  $I$                                   \\
$R$                                                   & the input report (a total of $M$   tokens)                                                          \\
$R_m$                                                 & masked report tokens (a total of $n$)                                                               \\
$R_u$                                                 & unmasked report tokens (a total of   $M-n$)                                                         \\
$E_t$                                                 & the report embedding obtained by   projecting $R_u$ into embedding space                            \\
$F(\cdot)$                                            & the proposed memory-augmented   cross-modal fusion module                                           \\
$C_v$, $C_{E_t}$                                      & the cross-modal intermediate   representation obtained by feeding $v$ and $E_t$ into MCA            \\
$M_v$, $M_{E_t}$                                      & learnable memory matrices in   the MA-CMF module                                                    \\
$C$                                                   & the concatenation of $C_v$ and   $C_{E_t}$                                                          \\
$t$                                                   & global report representations obtained   by projecting $R$ into the pre-trained text encoder        \\
$\Theta_{E_{I}}$, $\Theta_{D_{T}}$, $\Theta_{F}$ & the weight parameters of the shared   vision encoder, text decoder, and MA-CMF module, respectively  \\
\bottomrule
\end{tabular}
\label{tab:notations}
\end{table*}

\end{appendices}

\end{document}